%% file: paper.tex
\documentclass[]{fairmeta}

\usepackage[T1]{fontenc}

\usepackage[utf8]{inputenc}

\usepackage{microtype}

\usepackage{inconsolata}

\usepackage{graphicx}
\usepackage{lineno}

\usepackage{url}            %
\usepackage{booktabs}       %
\usepackage{xcolor}         %
\usepackage{adjustbox}
\usepackage{multirow}
\usepackage{amsmath}
\usepackage[most]{tcolorbox}
\usepackage{hyperref}
\usepackage{subcaption}
\usepackage{caption}
\usepackage{enumitem}
\usepackage[english]{babel}
\usepackage{float}
\usepackage{wrapfig}
\usepackage{xspace}
\usepackage{tablefootnote}

\newcommand{\dataname}{\textsc{NaturalThoughts}\xspace}
\title{\dataname: Selecting and Distilling Reasoning Traces for General Reasoning Tasks}

\author[*]{Yang Li}
\author[*]{Youssef Emad}
\author[*]{Karthik Padthe}
\author[*]{Jack Lanchantin}
\author{Weizhe Yuan}
\author{Thao Nguyen}

\author{\\Jason Weston}
\author{Shang-Wen Li}
\author{Dong Wang}
\author[\dagger]{Ilia Kulikov}
\author[\dagger]{Xian Li}

\affiliation{FAIR at Meta}

\contribution[*]{Joint first author}
\contribution[\dagger]{Joint last author}

\abstract{
Recent work has shown that distilling reasoning traces from a larger teacher model via supervised finetuning outperforms reinforcement learning with the smaller student model alone~\citep{guo2025deepseek}. However, there has not been a systematic study of what kind of reasoning demonstrations from the teacher are most effective in improving the student model's reasoning capabilities. 
In this work we curate high-quality \dataname{} by selecting reasoning traces from a strong teacher model based on  a large pool of questions from \textsc{NaturalReasoning}~\citep{yuan2025naturalreasoning}. 
We first conduct a systematic analysis of factors that affect distilling reasoning capabilities, in terms of sample efficiency and scalability for general reasoning tasks.  We observe that simply scaling up data size with random sampling  is a strong baseline with steady performance gains. Further, we find that selecting difficult examples that require more diverse reasoning strategies is more sample-efficient to transfer the teacher model's reasoning skills.
  Evaluated on both Llama and Qwen models, training with \dataname \space outperforms existing reasoning datasets such as OpenThoughts, LIMO, etc. on general STEM reasoning benchmarks including GPQA-Diamond, MMLU-Pro and SuperGPQA.

  }

\date{\today}
\correspondence{Xian Li at \email{xianl@meta.com}}

\begin{document}
\renewcommand{\sectionautorefname}{Section}
\renewcommand{\subsectionautorefname}{Section}
\renewcommand{\subsubsectionautorefname}{Section}

\maketitle
\input{meta_template_arxiv/sections/introduction}

\input{meta_template_arxiv/sections/related_work}

\input{meta_template_arxiv/sections/method}

\input{meta_template_arxiv/sections/experiments}

\input{meta_template_arxiv/sections/conclusion}

\bibliography{meta_template_arxiv/paper}
\bibliographystyle{meta_template_arxiv/assets/plainnat}

\input{meta_template_arxiv/sections/appendix}

\end{document}

%% file: meta_template_arxiv/sections/introduction.tex
\section{Introduction}

Scaling reinforcement learning (RL) with generated chain-of-thoughts (CoTs) has led to remarkable improvements in the reasoning capabilities of large language models (LLMs)~\citep{guo2025deepseek}.
An effective approach to elicit such capabilities, especially for smaller models, is to distill from a teacher model, i.e. supervised-finetuning (SFT) on reasoning traces output by a stronger reasoning model. In practice, SFT with thousands of training examples is often applied as a critical step before RL even for larger models~\citep{guo2025deepseek,bercovich2025llamanemotronefficientreasoningmodels}. The importance of distillation is also evident by the fact that RL alone does not increase the innate priors for reasoning a student  model~\citep{yue2025doesreinforcementlearningreally}, while SFT on reasoning traces from a teacher model 
can add new reasoning primitives to be explored in the RL stage.

Various efforts from the community have been devoted to distilling CoT trajectories from reasoning models, with notable examples being Open-R1~\citep{openr1} and OpenThoughts~\citep{guha2025openthoughts}. Through these efforts, it has been demonstrated that state-of-the-art reasoning models often exhibit sub-optimal reasoning trajectories such as ``overthinking'' or ``underthinking''~\citep{kumar2025overthink,chen2025think23overthinkingo1like,wang2025thoughtsplaceunderthinkingo1like}. Recent work such as LIMO \citep{ye2025limoreasoning} and S1K \citep{muennighoff2025s1simpletesttimescaling} show the importance of selecting and curating distillation examples to improve downstream capabilities. They find that as few as 1,000 high-quality reasoning traces are sufficient to drastically increase the student model's performance on mathematical reasoning tasks~\citep{ye2025limoreasoning,muennighoff2025s1simpletesttimescaling}. Both LIMO and S1K conduct manual selection of distillation examples, limited to math and coding questions, subsequently leading primarily to performance gains in those domains. While such curated small-scale datasets are helpful for solving easy to medium-difficulty problems in in-distribution domains, they may not generalize well to reasoning problems in other domains~\citep{sun2025climbingladderreasoningllms}. 

In this work, we conduct a systematic analysis of data-centric factors that affect distilling reasoning capabilities from a ``reasoning'' teacher model to an (initially) ``non-reasoning'' student model. Specifically, we use questions from NaturalReasoning as a testbed, given its diversity and effectiveness in eliciting reasoning~\citep{yuan2025naturalreasoning}, and generate reasoning traces from a performant reasoning model such as DeepSeek-R1~\citep{guo2025deepseek}. We use the resulting dataset, \dataname, to examine the effects of training on different filtered data subsets and understand how models learn to reason effectively. We study the curation of distillation training examples along several axes including scale, diversity, and difficulty. We find that increasing each axis in the training data leads to performance gains.

Given that an important consideration when deploying the student model in real-world applications is inference-time efficiency, we also take into account the susceptibility of current models to overthinking (i.e., generating too many reasoning tokens).
Consequently, in addition to reasoning performance, we also focus on improving the student model's reasoning \textit{efficiency}. We propose a simple training method with mixed System-1 and System-2 distillation \citep{yu2024distilling}, where training examples do not always contain the full reasoning traces from the teacher (System-2), but instead only contain the teacher model's final answers after thinking (System-1). %

\begin{figure*}[t]
    \centering
    
    \begin{subfigure}[b]{0.98\textwidth}%
        \includegraphics[width=\linewidth]{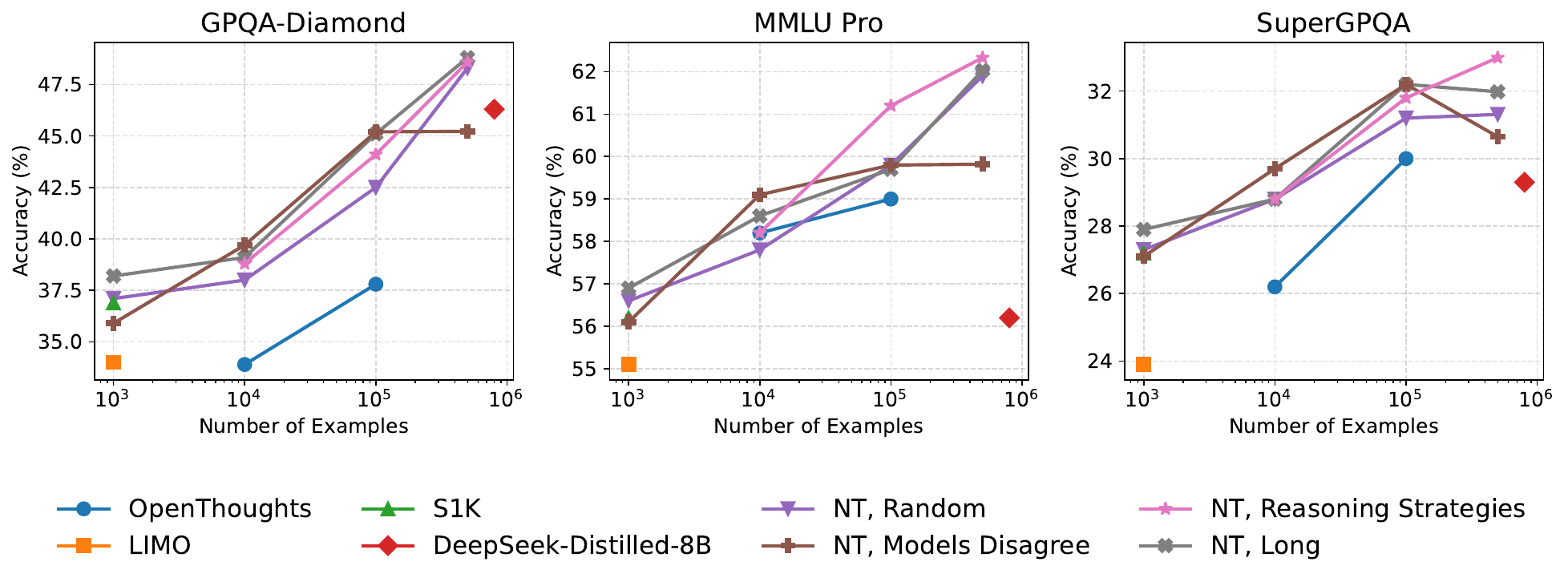}%
    \end{subfigure}
    \hfill %
   \begin{subfigure}[b]{0.98\textwidth}%
        \includegraphics[width=\linewidth]{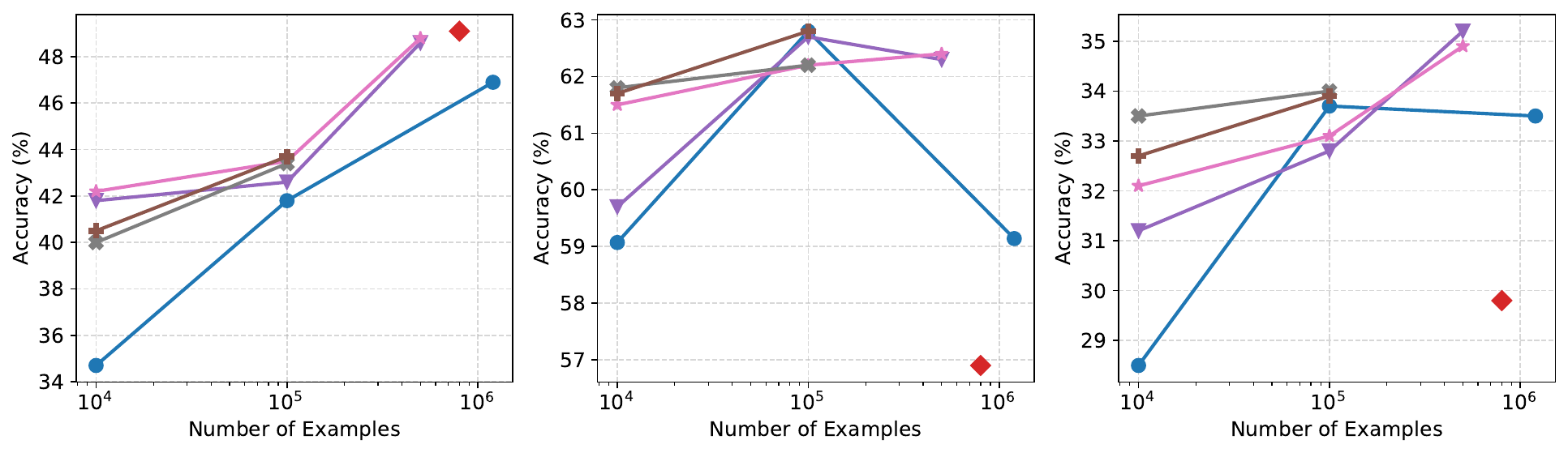}%
    \end{subfigure}
    \caption{Comparison of \dataname\space(NT) with existing distillation datasets, when training Llama-3.1-8B-Instruct (Top) and Qwen-2.5-7B-Instruct (Bottom) respectively. In contrast to the ``Less is More'' hypothesis~\citep{ye2025limoreasoning,muennighoff2025s1simpletesttimescaling}, we observe that scaling up high-quality questions and reasoning demonstrations consistently improves performance, even with random selection. Selection based on diversity in reasoning strategies and difficulty (e.g. long CoT, disagreement between teacher models) further improves over random selection (details in \autoref{subsec:selection} and \autoref{sec:data_selection}). }
    \label{fig:data-scaling}
\end{figure*}

Our main contributions are as follows:
\begin{itemize}[itemsep=1pt,topsep=0.5pt,leftmargin=13pt]
    \item We demonstrate the importance of \textbf{scaling} high-quality, diverse reasoning data, which is contrary to the ``Less is More'' hypothesis~\citep{ye2025limoreasoning,muennighoff2025s1simpletesttimescaling} (\autoref{sec:data_selection}). We observe consistent performance improvements by scaling up the data quantity, even with \textit{random} selection from \dataname.
    \item We systematically compare different data \textbf{selection} methods based on a large pool of reasoning traces from a teacher model (\autoref{sec:data_selection}). We examine the efficacy of selecting SFT examples based on various metrics including \textit{diversity} and \textit{difficulty}. We find that difficult examples which require longer reasoning chains and diverse reasoning strategies are most effective at distilling the teacher's reasoning capabilities. 
    \item To make the student model's reasoning process more \textbf{efficient}, we show that a simple method of training with mixed System-1 (using the condensed final responses) and System-2 (using both the intermediate CoTs and the final responses) reasoning based on question difficulty enables the student model to adapt its reasoning strategy based on the input. This allows for a dynamic trade-off between efficiency and accuracy at test time, resulting in more effective problem-solving and significantly shifting the efficiency-accuracy frontier.  (\autoref{sec:distillation_methods}).

\end{itemize}

%% file: meta_template_arxiv/sections/related_work.tex
\begin{figure*} %
    \centering
    \begin{minipage}[t]{0.485\textwidth} %
        \includegraphics[width=\linewidth]{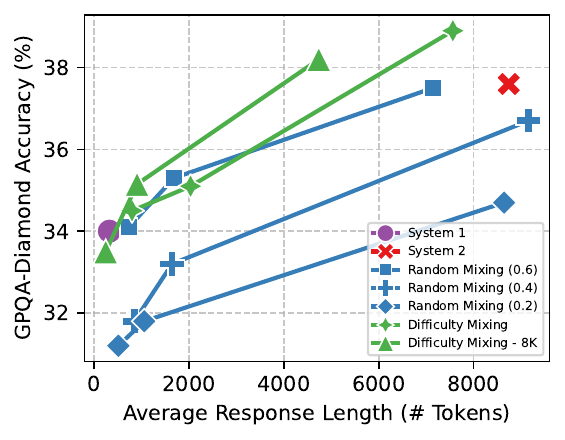}
        \caption{A mix of System-1 (no reasoning traces, final answer only) and System-2 (full reasoning traces) distillation improves the inference-time efficiency of the student model (details in \autoref{subsec:mix_reasoning} and \autoref{sec:distillation_methods}).}
        \label{fig:scaling_mix}
    \end{minipage}\hfill
    \begin{minipage}[t]{0.485\textwidth} %
        \includegraphics[width=\linewidth]{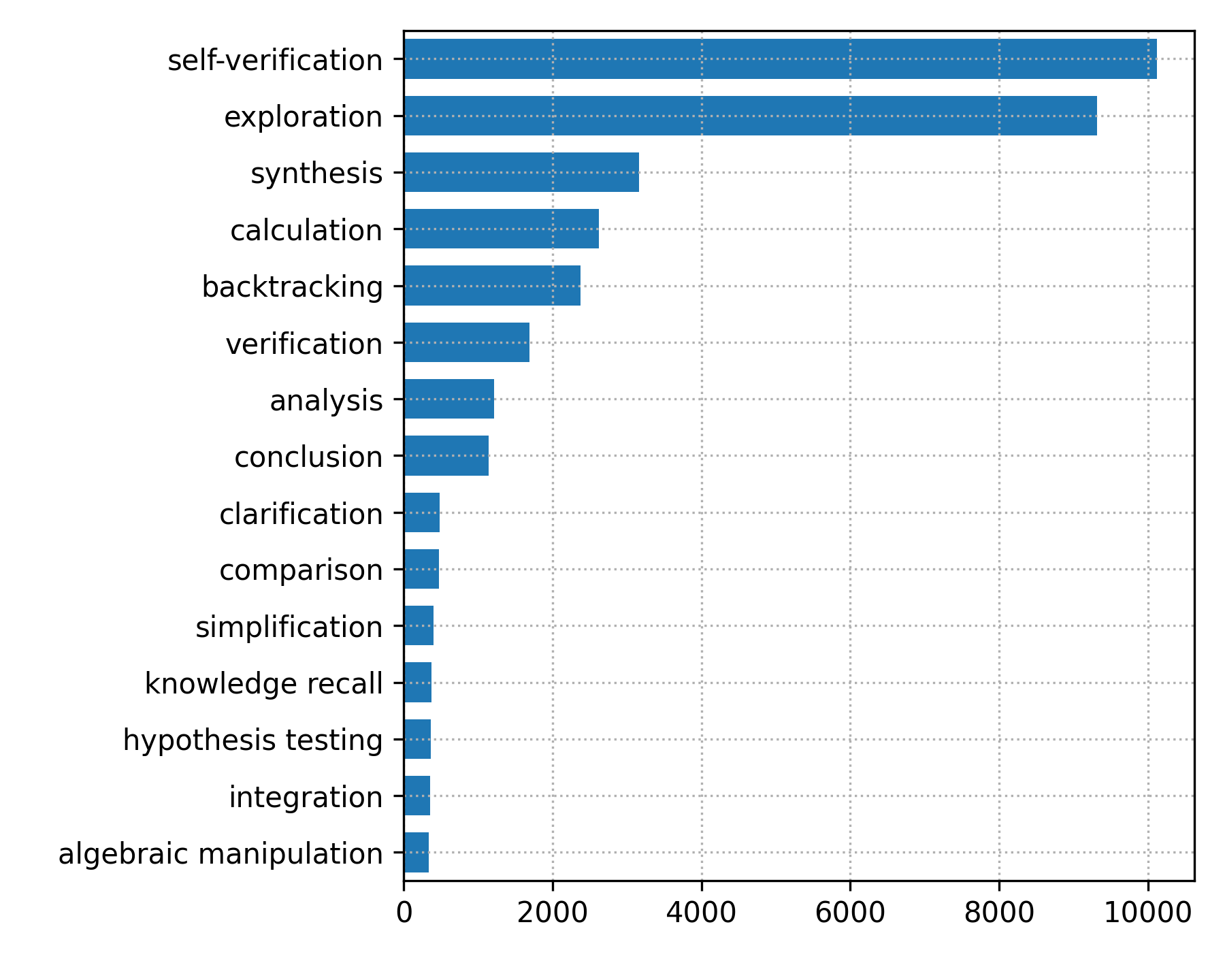}
        \caption{Top 20 commonly used reasoning strategies based on annotations of 10,000 samples. We observe that there are a small number frequently used strategies, followed by a long tail of more niche strategies.}
        \label{fig:reasoning_verbosity}
    \end{minipage}
\end{figure*}

\section{Related Work}
\textbf{Data-centric Approaches for Improving Reasoning\ } Distillation and reinforcement learning have become standard approaches for building strong reasoning models. ~\citet{guo2025deepseek} explicitly compare the effectiveness of distillation vs. RL for improving small models' reasoning capabilities, and find that distilling from a strong teacher model significantly outperforms large-scale RL training only on the student model. Since then, the community has been actively working on distillation and proposing distilled datasets such as OpenThoughts~\citep{guha2025openthoughts} and OpenR1~\citep{openr1}. However, there has been little study on the quality of distillation datasets until recently. Prior work such as LIMO~\citep{ye2025limoreasoning} and S1K~\citep{muennighoff2025s1simpletesttimescaling} shows that carefully curated questions and reasoning traces can greatly improve sample efficiency, where 1,000 examples are sufficient to distill the long CoT behaviors from the teacher model.
However, these hypotheses were primarily verified in narrow domains such as math and coding. Our work provides new insights on a much broader set of domains with more diverse reasoning problems. 

\textbf{Reasoning Efficiency\ } Recent work has shown that overthinking is a common pattern in SoTA reasoning models (e.g., R1~\citep{guo2025deepseek}, o1~\citep{jaech2024openai}, o3, etc.) and thus the need to curate high-quality reasoning traces~\citep{kumar2025overthink,chen2025think23overthinkingo1like} as well as to develop methods that elicit more efficient reasoning~\citep{aggarwal2025l1controllinglongreasoning}. In particular, ~\citet{ma2025reasoningmodelseffectivethinking} demonstrate non-trivial reasoning performance by simply omitting thinking in the training data.  An alternative approach to selecting reasoning traces is to revise the teacher model's reasoning traces and generate new reasoning paths ~\citep{lu2025retrosearchexploringuntakenpaths}. Distilling System 2 into System 1 was first proposed in ~\citet{yu2024distilling21}.

%% file: meta_template_arxiv/sections/method.tex
\section{Method}
\begin{figure}
    \centering
    \fbox{\includegraphics[width=1.0\linewidth]{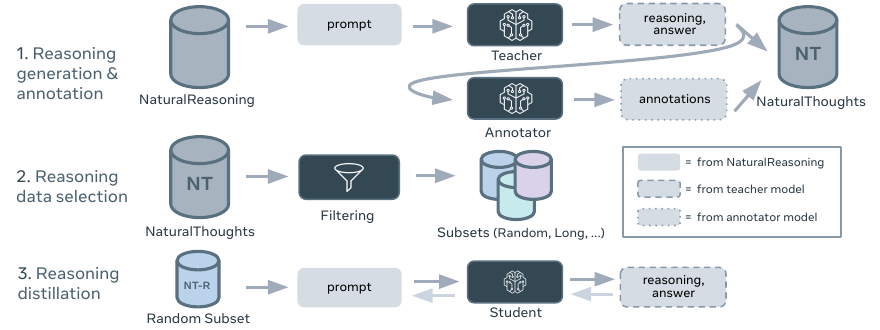}}
    \caption{\dataname{} Overview. We outline the three main components of our strategies for selecting and distilling reasoning data. 1) Starting from the NaturalReasoning dataset \citep{yuan2025naturalreasoning}, we sample prompts for the teacher model to generate reasoning and answer traces from, creating the \dataname{} dataset. 2) From the \dataname{} seed set, we use several filtering methods for data selection (\autoref{sec:data_selection}). 3) Given a filtered subset obtained from data selection, we finetune a student model on the prompt-reasoning-answer triples, where the reasoning traces and answers are generated by the teacher model.}
    \label{fig:enter-label}
\end{figure}

\label{sec:method}
The objective of our work is to distill reasoning capabilities from a ``reasoning'' teacher model to an (initially) ``non-reasoning'' student model.
We define each reasoning training example to be a (question, reasoning response) pair, where the response is generated by a teacher model, consisting of two parts: the intermediate reasoning trace (e.g. the tokens between \texttt{<think>} and \texttt{</think>}, which represent \textit{System-2} reasoning) and the final answer (\textit{System-1}). We sample questions from NaturalReasoning ~\citep{yuan2025naturalreasoning}, a comprehensive dataset comprising 2.8 million questions that span multiple domains and that have been shown to be effective in eliciting reasoning. We use DeepSeek-R1 ~\citep{guo2025deepseek} as the teacher model to generate reasoning responses. We call the resulting dataset of distilled reasoning examples \dataname. In the following subsections, we outline how we annotate and filter reasoning examples from the initial dataset in order to optimize the student model's reasoning capabilities.

\subsection{Reasoning Annotation}
\label{sec:annotation}
Given a training set of (question, reasoning response) pairs, we first seek to annotate the data, using three different annotations.
First, we annotate the domain and topics of each question using the taxonomy from ~\citet{du2025supergpqa}, which includes 13 top-level domains including Engineering, Philosophy, Medicine, Economics, Science, Law, History, Education, Management, Literature and Arts, Agronomy, Sociology and Military Science. 
Then, for each reasoning trace, we prompt Llama-3.1-70B-Instruct~\citep{grattafiori2024llama3herdmodels} to identify the ``meta-reasoning'' strategies throughout the thinking process, such as self-verification, backtracking, exploration, etc. 
Finally, we prompt Llama-3.1-70B-Instruct to score the ``verbosity'' of the reasoning, from 0 to 10, where 0 means the reasoning is very efficient with no rambling, and 10 means excessive rambling and not making progress towards a solution. The full prompts used for annotations are provided in \autoref{tab:prompt_annotate_reasoning} and ~\autoref{tab:prompt_annotate_domain_5}.

\subsection{Reasoning Data Selection}
\label{subsec:selection}

Given the training examples with their respective annotations, the second step is to select samples used for distillation. We study different data selection strategies along the following two axes: diversity and difficulty.

\textbf{Diversity\ } We hypothesize that a diversified set of questions and reasoning traces would be effective for distillation~\citep{muennighoff2025s1simpletesttimescaling}. We attempt to obtain diverse subsets of data using three properties.
\begin{itemize}[itemsep=1pt,topsep=1pt,leftmargin=13pt]
    \item \textbf{Question topics} With annotated question topics from~\autoref{sec:annotation}, we sample data uniformly across all the 12 topic domains, taking about 850 samples from each domain. 

    \item \textbf{Question semantic embeddings} We also create a diverse subset of questions using the semantic feature space. Specifically, we embed questions using Llama 3.1-8B-Instruct. We then perform density-based clustering on the embeddings, and uniformly sample from each cluster. More details can be found in \autoref{app:clustering}.

    \item \textbf{Reasoning strategies} With the annotated reasoning strategies from \autoref{sec:annotation}, each example has a set of strategies $\mathcal{S}= \{s_i\}$.  As shown in ~\autoref{fig:reasoning_verbosity}, there are a few frequently used strategies such as self-verification, etc. followed by a long tail of niche strategies. To select examples demonstrating diverse problem-solving strategies yet without ``overthinking'', we downsample examples where the number of reasoning strategies $|S| \le R_{min}$ or $|S| > R_{max}$. In the experiments in \autoref{sec:data_selection}, we use $R_{min}=4$ and $R_{max}=8$, based on the distribution of unique strategies annotated per example (\autoref{fig:unique_reasoning_hist}). We also downsample examples with low reasoning density, measured as having fewer unique reasoning strategies than the number of reasoning steps. Further ablations on the sampling approach are provided in \autoref{app:reasoning_strategies}.

\end{itemize}

\textbf{Difficulty\ } Another hypothesis is that the quality of the reasoning traces is correlated with the difficulty of the questions, which usually requires advanced reasoning. We therefore attempt to create data subsets with varying levels of difficulty using the following strategies.
\begin{itemize}[itemsep=1pt,topsep=1pt,leftmargin=13pt]
    \item \textbf{Length}
    The initial training dataset is dominated by short traces. Therefore, we downsample examples with short reasoning responses to study the benefit of using longer reasoning chains. Specifically, each example is sampled with probability
$p=(l/{C})^{\tau}$, where $l$ is the length of reasoning response measured by the number of tokens, C is a constant normalizer, and $\tau$ is the sampling temperature modulating how heavily shorter sequences are downsampled. In the experiments in \autoref{sec:data_selection}, we use $C=5000$, $\tau=2.5$.
    \item \textbf{Verbosity} Given the annotated verbosity scores from ~\autoref{sec:annotation}, we derive three subsets by sampling without replacement based on the scores: Low (beginning with the lowest verbosity, 0, and progressively including samples with higher verbosity), High (starting from the highest verbosity, 10, and progressively including samples with lower verbosity), and Med (including all samples with a verbosity of 5).
    \item \textbf{Models Agreement} For each example, we compared responses from a model with long CoT reasoning traces (Deepseek-R1) and a model without long CoT traces (Llama-3.3-70B). We use their disagreement, judged by Llama-3.1-8B-Instruct, as a proxy of question difficulty. We create  two subsets of training examples based on solution agreement or disagreement.
\end{itemize}

\subsection{Mixed Reasoning Distillation}
\label{subsec:mix_reasoning}
As the teacher reasoning model may have sub-optimal reasoning patterns such as ``overthinking'' or ``underthinking'', we compare different settings of distilling the teacher model's reasoning:
    
    \textbf{System-2 Distillation\ } By default, we conduct supervised finetuning on the entire response generated by the teacher model, which includes the full reasoning trace and the final answer.
    
    \textbf{System-1 Distillation\ } Instead of learning from the long CoT reasoning trace, we investigate the effectiveness of only learning from the teacher's final answer.
    
    \textbf{Mixed System-1  and System-2\ } Training data is a mixture of examples from both types as described above. We compared two mixing approaches: \textit{random mixing} and \textit{difficulty-based mixing}. In random mixing, we select training examples with full System-2 reasoning with probability $p=\{0.2, 0.4, 0.6\}$. In difficulty-based mixing, we use full System-2 reasoning traces for examples annotated with disagreement (as a proxy of difficult questions) and only use the condensed System-1 response for the remaining examples.

\textbf{Adaptive Reasoning at Inference-time\ } To enable explicit control of which types of reasoning to use at inference time, we append an explicit instruction to the end of the question to indicate which types of reasoning and how much inference budget the response should use. Specifically, we augment the training data with ``\texttt{Think carefully before answering. Use about \{K\} words.}'' for System-2, and ``\texttt{Answer directly without thinking. Use about \{K\} words.}'' for System-1. \texttt{K} is derived from the training data. At inference time, we evaluate the accuracy-efficiency trade-offs under three settings:

\begin{itemize}[itemsep=1pt,topsep=2pt,leftmargin=13pt]
  \item \textbf{No-Think} We instruct the model to  ``\texttt{Answer directly without thinking}'', i.e. perform System-1 mode of reasoning by generating a short condensed answers.
  \item \textbf{Think} We instruct the model to ``\texttt{Think carefully before answering. Use about \{K\} words}'' followed by the special begin-of-reason token \texttt{<think>} for force the generation into a full System-2 mode. $K$ is set to 3500 which is the average length of System-2 responses in the training set.
  \item \textbf{Adaptive-Think} To test whether the mixed System-1 and System-2 distillation can enable the student model to efficiently and automatically adapt to the question difficulty at inference time, we also evaluate a \textit{hybrid} mode, where we instruct the model to ``Think carefully before answering.'' but \textit{without} explicitly appending the special token \texttt{<think>}.
\end{itemize}

%% file: meta_template_arxiv/sections/experiments.tex
\section{Experimental Setup}
\label{sec:experiment}
We perform supervised finetuning (SFT) with \dataname{} data on both Llama-3.1-8B-Instruct and Qwen2.5-7B-Instruct student models. We compare different data selection methods at the scale of 1,000 and 10,000 training examples. We also experiment with training on 100,000 and 500,000 samples to test the scaling properties of \dataname. During SFT training, we mask the loss on the question, i.e. only train on the teacher model's reasoning trace and final answer. Unless otherwise noted, we use a maximum response length of 16,384 tokens and ensure that each training example contains a complete response within the maximum number of tokens. We utilize the fairseq2 library \citep{balioglu2023fairseq2} for training. We apply dynamic batching where each batch roughly contains 400k tokens. We train 10 epochs for 1k samples, 6 epochs for 10k, and 8 epochs for 100k and 500k training examples respectively. We use the AdamW optimizer with 0.1 weight decay and a constant learning rate $2e-5$. Each model is trained on 32 NVIDIA H200 GPUs.

\textbf{Baselines\ } We compare the curation approaches described in \autoref{subsec:selection} with the following baselines:  
\begin{itemize}[itemsep=1pt,topsep=1pt,leftmargin=13pt]
    \item \textbf{Random selection} We randomly select 1k, 10k, 100k and 500k examples from \dataname{} (\textsc{NT-Random}).
    \item \textbf{State-of-the-art reasoning datasets} We compare to \textsc{LIMO}~\citep{ye2025limoreasoning} and \textsc{S1K}~\citep{muennighoff2025s1simpletesttimescaling} which use the same teacher model (DeepSeek-R1) and perform careful manual curation for both the questions and the reasoning responses. For a fair comparison, we train our own Llama-3.1-8B-Instruct student model on their datasets.  To understand the effect of question source and question quality in \dataname, we also build another two baselines (\textsc{NT-NN-LIMO}, \textsc{NT-NN-S1K}), which use questions from \textsc{LIMO} and \textsc{S1K} as seeds to retrieve similar questions from \dataname. Specifically,  we choose the nearest neighbors of the LIMO and S1K questions in the question embedding space, and train a Llama-3.1-8B-Instruct student model on the retrieved data. We also compare to \textsc{OpenThoughts3}~\citep{guha2025openthoughts}, a high-quality reasoning dataset that has achieved state-of-the-art performance on several reasoning benchmarks. We compare \dataname{} to \textsc{OpenThoughts3} using both Llama-3.1-8B-Instruct and Qwen2.5-7B-Instruct student models\footnote{Specifically, we compare to different released assets from \textsc{OpenThoughts3} for different settings. For experiments with Llama models, as described in \autoref{tab:selection-8b}, we compare to \url{https://huggingface.co/mlfoundations-dev/openthoughts3_10k_llama3} and \url{https://huggingface.co/mlfoundations-dev/openthoughts3_100k_llama3}. For experiments with Qwen models, as described in \autoref{tab:selection-qwen}, we compare to \url{https://huggingface.co/mlfoundations-dev/openthoughts3_10k}, \url{https://huggingface.co/mlfoundations-dev/openthoughts3_100k} and \url{https://huggingface.co/open-thoughts/OpenThinker3-7B}.}.
    \item \textbf{DeepSeek-R1 distilled models} As a reference point, we also compare to the DeepSeek-R1-Distill-Llama-8B and DeepSeek-R1-Distill-Qwen-7B \citep{guo2025deepseek} models, which were trained via SFT with the same teacher and student models as our experiments, but with 800k non-public data.
\end{itemize}

\textbf{Evaluation\ }
We evaluate on reasoning benchmarks in both math and general STEM domains: MATH-500~\citep{hendrycks2021measuring} (500 examples), GPQA-Diamond~\citep{rein2024gpqa} (198 examples), MMLU-Pro~\citep{wang2024mmlu} (12,032 examples), and SuperGPQA~\citep{du2025supergpqa} (26,529 examples). We evaluate all models using the same evaluation setting as used in \citet{guo2025deepseek}. Specifically, we report $pass@1$ averaged across multiple seeds to reduce the variance. We use $N=24$ seeds for GPQA-Diamond, and $N=16$ for MATH-500, and N=1 for MMLU-Pro and SuperGPQA. For decoding hyperparameters, we use temperature $0.6$, and top-p $0.95$ for Llama models, and temperature $0.7$, top-p $1.0$ for Qwen models. We use maximum generation length of 16,384 tokens to match the sequence lengths used in training.

\section{Results}
\begin{table*}[!ht]
    \centering
    \small
        
        \vspace{0.2cm}
      \renewcommand{\arraystretch}{0.9} %
      \begin{adjustbox}{max width=1.0\textwidth}
        \begin{tabular}{lccccc}
            \toprule
            & Train Size & GPQA-D & MATH-500 & MMLU-Pro & SuperGPQA \\
            \midrule
            \textit{Baselines} & & & & \\
            \hspace{3mm}Llama-3.1-8B-Instruct & 0 & $29.0_{\pm0.5}$ & $49.1_{\pm0.4}$ & 47.7 & 21.9  \\
            \hspace{3mm}LIMO~\citep{ye2025limoreasoning}  &817 & $34.0_{\pm0.5}$ & $56.5_{\pm0.3}$ & $55.1$ & $23.9$ \\
            \hspace{3mm}S1K~\citep{muennighoff2025s1simpletesttimescaling}   &1k & $36.9_{\pm0.7}$ & $59.4_{\pm0.3}$ & $56.2$ & $27.2$ \\
            \hspace{3mm}NT-NN-LIMO & 10k & $35.3_{\pm0.6}$ & $64.7_{\pm0.4}$ & 57.6 & 27.7\\
            \hspace{3mm}NT-NN-S1K & 10k & $38.5_{\pm0.6}$ & $63.7_{\pm0.3}$ & 57.4 & 29.0\\
            \hspace{3mm}OpenThoughts3~\citep{guha2025openthoughts} & 10k &  $33.9_{\pm0.5}$ & $72.3_{\pm0.4}$ & 58.2 & 26.2  \\
            \hspace{3mm}OpenThoughts3~\citep{guha2025openthoughts} & 100k &  $37.8_{\pm0.5}$ & $82.2_{\pm0.2}$ & 59.0 & 30.0 \\
            \hspace{3mm}DeepSeek-R1-Distill-Llama-8B & 800k & 46.3 & 89.1 & 56.2 & 29.3 \\
            \midrule
            \dataname,\textit{ Scale} & & & & \\
            \hspace{3mm}Random & 1k & $37.1_{\pm0.4}$ & $57.8_{\pm0.3}$ & 56.6 & 27.3  \\
            \hspace{3mm}Random & 10k &  $37.6_{\pm0.6}$ & $61.3_{\pm0.3}$ & 57.8 & 28.8  \\
            \hspace{3mm}Random & 100k &  $42.5_{\pm0.6}$ & $67.5_{\pm0.3}$  & 59.8 & 31.2  \\
            \hspace{3mm}Reasoning Strategies  & 100k& $44.1_{\pm0.6}$ & $67.7_{\pm0.3}$ & 61.2 & 31.8\\
            \hspace{3mm}Models Disagree  & 100k& $45.2_{\pm0.6}$ & $70.2_{\pm0.2}$ & 59.8 & 32.2\\
            \hspace{3mm}Long & 100k & $43.1_{\pm0.7}$ & $69.0_{\pm0.3}$ & 61.2 & 32.2\\
            \hspace{3mm}Random & 500k &  $48.3_{\pm0.6}$ & $72.3_{\pm0.3}$ & 61.9 & 31.3  \\
            \hspace{3mm}Reasoning Strategies & 500k &  $48.6_{\pm0.7}$ & $75.4_{\pm0.2}$ & 62.3 & 33.0  \\
            \hspace{3mm}Models Disagree & 500k &  $45.2_{\pm0.6}$ & $70.8_{\pm0.3}$ & 59.8 & 30.7  \\
            \hspace{3mm}Long & 500k &  $48.8_{\pm0.6}$ & $74.9_{\pm0.3}$ & 62.0 & 31.9  \\
            
            \midrule
            \dataname,\textit{ Diversity} & & & & \\
            \hspace{3mm}Topics & 10k& $32.7_{\pm0.5}$ & $55.7_{\pm0.3}$ & 55.9 & 25.4\\
            \hspace{3mm}Semantic Embeddings& 10k & $39.4_{\pm0.5}$ & $60.3_{\pm0.3}$ & 55.9 & 27.2 \\
            \hspace{3mm}Reasoning Strategies& 10k & $38.8_{\pm0.5}$ & $63.5_{\pm0.4}$ & 58.2 & 28.8\\
            \midrule
            \dataname,\textit{ Difficulty} & & & & \\
            \hspace{3mm}Long & 10k& $39.1_{\pm0.7}$ & $63.5_{\pm0.3}$ & 58.6 & 28.8\\
            \hspace{3mm}Models Disagree    & 10k& $39.7_{\pm0.4}$ & $61.9_{\pm0.3}$ & 59.1 & 29.7\\
            \hspace{3mm}Models Agree  & 10k& $37.5_{\pm0.6}$ & $60.1_{\pm0.3}$ & 56.7 & 28.9\\
            \hspace{3mm}Verbosity=Low  & 10k  & $37.0_{\pm0.7}$ & $59.1_{\pm0.3}$ & 54.8 & 26.7\\
            \hspace{3mm}Verbosity=Med  & 10k  & $37.2_{\pm0.6}$ & $62.1_{\pm0.4}$ & 57.5 & 28.9\\
            \hspace{3mm}Verbosity=High & 10k & $38.4_{\pm0.6}$ & $59.8_{\pm0.4}$ & 56.6 & 28.7\\

            \bottomrule
        \end{tabular}
        \end{adjustbox}
        \caption{\textbf{Reasoning data scaling and selection for the Llama-3.1-8b-Instruct student model.} We compare reasoning data selection based on quality, diversity and difficulty. We use DeepSeek R1 as the teacher model, and conduct supervised finetuning on Llama-3.1-8B-Instruct as the student model. Selection based on diversity in reasoning strategies and question difficulty (via model disagreement) outperforms random selection, although the gap becomes smaller when scaling up data size. 
        }
        \label{tab:selection-8b}
        \vspace{-0.40cm}
\end{table*}

\begin{table*}[!h]
    \centering
    \small
        
        \vspace{0.2cm}
      \renewcommand{\arraystretch}{0.9} %
      \begin{adjustbox}{max width=1.0\textwidth}
        \begin{tabular}{lccccc}
            \toprule
            & Train Size & GPQA-D & MATH-500 & MMLU-Pro & SuperGPQA \\
            \midrule
            \textit{Baselines} & & & & \\
            \hspace{3mm}Qwen-2.5-7B-Instruct & 0 & $34.1_{\pm0.5}$ & $69.7_{\pm0.3}$ & 56.5 & 29.5  \\
            \hspace{3mm}OpenThoughts3~\citep{guha2025openthoughts} & 10k & $34.7_{\pm0.5}$ & $83.2_{\pm0.2}$ & 59.1 & 28.5\\
            \hspace{3mm}OpenThoughts3~\citep{guha2025openthoughts} & 100k & $41.8_{\pm0.5}$ & $88.1_{\pm0.2}$ & 62.8 & 33.7\\
            \hspace{3mm}OpenThoughts3~\citep{guha2025openthoughts} & 1.2m &  $46.9_{\pm0.5}$ & $91.2_{\pm0.1}$ & 59.1 & 33.5  \\
            \hspace{3mm}DeepSeek-R1-Distill-Qwen-7B & 800k & 49.1 & 92.8 & 56.8 & 29.8 \\ %
            \midrule
            \dataname & & & & \\
            \hspace{3mm}Random & 10k &  $41.8_{\pm0.6}$ & $79.0_{\pm0.2}$ & 59.7 & 31.2  \\

            \hspace{3mm}Reasoning Strategies & 10k & $42.2_{\pm 0.6}$ & $78.6_{\pm 0.2}$ & 61.5 & 32.1\\

            \hspace{3mm}Models Disagree   & 10k& $40.5_{\pm0.5}$ & $78.4_{\pm0.3}$ & 61.7 & 32.7\\
            \hspace{3mm}Long & 10k &  $40.0_{\pm 0.6}$ & $78.8_{\pm 0.2}$ & 61.8 & 33.5\\

            \hspace{3mm}Random   & 100k& $42.6_{\pm 0.5}$ & $80.8_{\pm 0.2}$ & 62.7 & 32.8 \\
            \hspace{3mm}Models Disagree   & 100k& $43.7_{\pm0.6}$ & $80.8_{\pm0.2}$ & 62.8 & 33.9\\
            \hspace{3mm}Reasoning Strategies   & 100k & $43.5_{\pm 0.5}$ & $80.9_{\pm 0.2}$ & 62.2 & 33.1 \\
            \hspace{3mm}Long & 100k & $43.4_{\pm 0.4}$ & $81.7_{\pm 0.4}$ & 62.2 & 34.0 \\
            \hspace{3mm}Random & 500k & $48.6_{\pm 0.6}$ & $83.1_{\pm 0.2}$ & 62.3 & 35.2 \\ %
            \hspace{3mm}Reasoning Strategies & 500k & $48.3_{\pm 0.5}$ & $83.6_{\pm 0.2}$ & 62.7 & 35.2 \\ %
            
            \bottomrule
        \end{tabular}
        \end{adjustbox}
        \caption{\textbf{Reasoning data scaling and selection for the Qwen-2.5-7B-Instruct student model.} We also provide comparisons when using Qwen-2.5-7B-Instruct as the student model. Training with 500k examples from \dataname{} outperforms training with 1.2m examples from  OpenThoughts3~\citep{guha2025openthoughts} on three of the four evaluation benchmarks. 
        }
        \label{tab:selection-qwen}
\end{table*}

\subsection{Reasoning Data Selection}
\label{sec:data_selection}
\autoref{tab:selection-8b} and \autoref{tab:selection-qwen} show the overall results of the different curation strategies for reasoning distillation from the teacher model, using Llama and Qwen student models respectively. We describe our findings in more details in the following paragraphs.

\textbf{Baseline Comparisons\ } First we obtain similar observations as in \textsc{LIMO}~\citep{ye2025limoreasoning} and \textsc{S1K}~\citep{muennighoff2025s1simpletesttimescaling}: with carefully selected training samples and a strong teacher model, we can effectively distill strong reasoning capabilities even at small data scales. As shown in ~\autoref{tab:selection-8b}, using only 1,000 \textit{randomly selected} examples from  \dataname already outperforms \textsc{LIMO} and is on par with \textsc{S1K}, even though these two prior datasets went through rigorous manual selection of questions and reasoning traces.

\textbf{Diversity\ }
In \autoref{tab:selection-8b}, we compare selecting examples based on diversity in \textit{questions} (both topics and semantic embeddings), which is often used in prior work ~\citep{muennighoff2025s1simpletesttimescaling}, as well as our proposed diversity in \textit{reasoning strategies}. First, we find that simply uniformly sampling across question topics reduces the performance compared to randomly sampling from the seed set. This is likely due to question topic distribution being too concentrated among a small set of topics. Compared to question topic diversity, the question semantic embedding diversity method produces better performance as the clusters are more fine-grained. Notably, we find that selecting samples based on the diverse \textit{reasoning strategies} results in the best performance out of the three selection criteria. This implies that the diversity of the reasoning traces is more important than the diversity of the questions themselves.

\textbf{Difficulty\ } 
We report multiple findings based on training on subsets of varying difficulty. First, the ``Long'' subset, or examples where shorter reasoning traces are downsampled, performs better than random selection which contains more short reasoning traces. Second, the ``Models Disagree'' subset performs better than both random selection and ``Long''. This implies that training on difficult questions that require more advanced reasoning traces is more sample-efficient in distilling reasoning capabilities. Overall, we find that the ``Models Disagree'' subset leads to the best average performance across all 10K filtered subsets.

Comparing the effects of reasoning verbosity, we find that the ``Medium Verbosity'' subset is on average the best performing verbosity training set. We hypothesize that the high verbosity subset has too many incoherent reasoning traces, and the low verbosity subset does not contain enough useful reasoning.
However, the verbosity distribution is quite concentrated, so it may not be a strong signal to distinguish high-quality examples.

\subsection{Reasoning Data Scaling}
\label{sec:data_scaling}

Different from what was observed in \textsc{LIMO} and \textsc{S1K}, where increasing the number of training samples results in marginal improvement or even decrease in distillation performance, we observe that with \dataname{}, simply scaling up data size beyond 1,000 to 10,000 improves performance on tasks requiring both knowledge and reasoning. This holds for both Llama models, as is shown in \autoref{tab:selection-8b} (\textit{scale}) and Qwen models, as is shown in \autoref{tab:selection-qwen}, even with random selection. Additionally, using manually curated data as question seeds (\textsc{NT-NN-LIMO}, and \textsc{NT-NN-S1K}) improves the scaling trend compared to random selection. We find that performance across the diverse set of evaluation tasks does not saturate when scaling up the training set size to 500,000 examples, with both random selection and selection based on reasoning strategy diversity (see \autoref{fig:data-scaling}).

We also verify whether the scaling trends persist with a larger student model, which is often more sample efficient. To this end, we use the Llama-3.3-70B-Instruct as the student model. We randomly select 1,000, 10,000 and 100,000 questions from \dataname. Results are summarized in \autoref{tab:70B_scaling}. Similar to the trend observed with Llama-3.1-8B-Instruct and Qwen-2.5-7B-Instruct, scaling up data size beyond 1,000 examples consistently improves performance across all four tasks. With 100k training samples, the student model already outperforms DeepSeek-R1-Distill-Llama-70B on GQPA-D, MMLU-Pro and SuperGPQA.   

\begin{table*}[h]
    \centering
    \small
    \renewcommand{\arraystretch}{0.9} %
    
    \begin{adjustbox}{max width=0.95\linewidth}
        \begin{tabular}{lccccc}
            \toprule
            & Train Size &  GPQA-D & MATH-500 & MMLU-Pro & SuperGPQA \\
            \midrule
            \dataname & & & & \\
            \hspace{3mm}Random Selection & 1k & $63.5_{\pm0.6}$ & $86.3_{\pm0.2}$ & 78.0 & 46.0\\
            \hspace{3mm}Random Selection& 10k & $65.6_{\pm0.6}$ & $87.4_{\pm0.3}$ & 78.4 & 48.3 \\
            \hspace{3mm}Random Selection& 100k & $67.6_{\pm0.4}$ & $88.5_{\pm0.2}$ & 78.9 & 50.6 \\
            \midrule
            DeepSeek-R1-Distill-Llama-70B & 800k & 65.2 & 94.5 & 78.5 & 49.4 \\
            \bottomrule
        \end{tabular}
    \end{adjustbox}
    \caption{\textbf{Scaling results with a larger model.} On general STEM reasoning tasks, training Llama3.3-70B-Instruct with \dataname  outperforms DeepSeek-R1-Distill-Llama-70B, which was trained with the same teacher and student models using more data.}
    \vspace{0.2cm}
    \label{tab:70B_scaling}
    \vspace{-0.10cm}
\end{table*}

\subsection{Mixed Distillation Efficiency}
\label{sec:distillation_methods}
We compare different reasoning distillation approaches outlined in \autoref{subsec:mix_reasoning}, and evaluate on GPQA-Diamond. We summarize results in \autoref{tab:mixed_think}.

\begin{table*}[!t]
    \centering
    \small
        
        \vspace{0.2cm}
      \renewcommand{\arraystretch}{0.9} %
      \begin{adjustbox}{max width=1.0\textwidth}
        \begin{tabular}{lccccccc}
            \toprule

            \ &\multicolumn{2}{c}{\textbf{No-Think}} &\multicolumn{2}{c}{\textbf{Adaptive-Think}} &\multicolumn{2}{c}{\textbf{Think}} 
            \\
            \cmidrule(lr){2-3} \cmidrule(lr){4-5} \cmidrule(lr){6-7}
            & Mean Length  & Accuracy & Mean Length  & Accuracy &  Mean Length  & Accuracy  \\
            \midrule
            System-1   & 321.3 & $34.0_{\pm0.6}$ & 541.1  & $32.4_{\pm0.7}$ & - & -\\
            \midrule
            System-2   & 5133.7 & $36.3_{\pm0.5}$ & 7817 & $37.3_{\pm0.6}$ & 8740.6 &$37.6_{\pm0.6}$ \\
            \midrule
            \textit{Random Mixing} & & & & \\
            \hspace{3mm}$p_{System2}=0.2$  &511.0 &$31.2_{\pm0.7}$& 1056.0 &$31.8_{\pm0.6}$  & 8642.3 & $34.7_{\pm0.7}$\\
            \hspace{3mm}$p_{System2}=0.4$  &847.3 &$31.8_{\pm0.5}$& 1642.7 &$33.2_{\pm0.5}$  & 9155.0  & $36.7_{\pm0.6}$ \\
            \hspace{3mm}$p_{System2}=0.6$  &729.6 &$34.1_{\pm0.6}$ &1683.4 &$35.3_{\pm0.6}$  &7139.8  & $37.5_{\pm0.5}$ \\
            \midrule
            Difficulty-based Mixing &799.4 &$34.5_{\pm0.5}$  &2033.0 &$35.1_{\pm0.6}$ &7562.4 & $38.9_{\pm0.7}$ \\
            \bottomrule
        \end{tabular}
        \end{adjustbox}
        \caption{\textbf{Mixed Reasoning Distillation.} We show the accuracy-efficiency tradeoff of training with mixed System-1 and System-2 reasoning, with three modes at inference time: ``No-Think'', ``Adaptive-Think'', and ``Think''. We report pass@1 accuracy on GPQA-Diamond. The mixed training approach enables operation in all three inference modes, which was not possible with either System-1 or System-2 alone. Our results demonstrate that difficulty-based mixing not only achieves a favorable accuracy-efficiency tradeoff, but also improves overall accuracy in ``Think'' mode compared to only using System-2. See Figure \ref{fig:scaling_mix} for a visualization of the trade-offs.
}
        \label{tab:mixed_think}
        \vspace{-0.5em}
\end{table*}

\begin{figure}
    \centering
    \subfloat{{\includegraphics[width=0.4\linewidth]{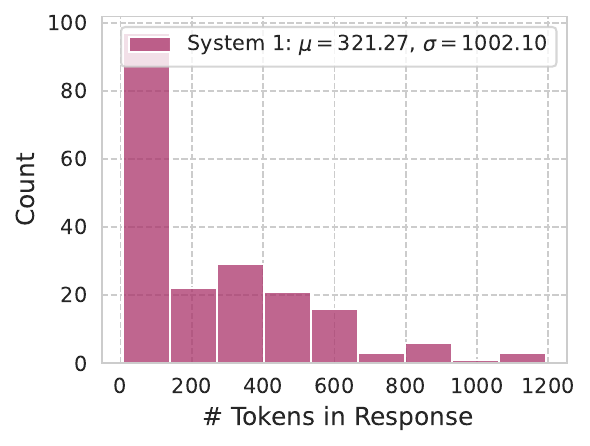} }}%
    \subfloat{{\includegraphics[width=0.4\linewidth]{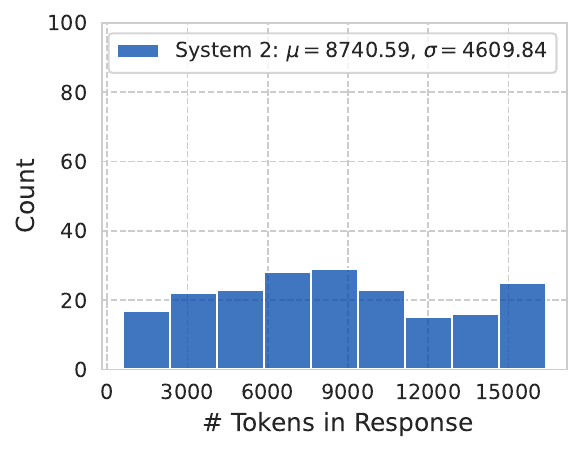} }}%
    \caption{\textbf{Generation length distributions of System-1 and System-2 reasoning for GPQA-Diamond.} System-1 responses (left) are predominantly brief, indicating minimal thinking, if any. In contrast, System-2 responses (right) are significantly longer than those of System-1 and exhibit greater variance, as the response lengths vary based on the amount of thinking required for each question, depending on its complexity.}
    \label{fig:sys1_sys2}
\vspace{-0.5em}
\end{figure}

\textbf{System-2 Distillation\ }
The baseline System-2 training (i.e. leveraging both the CoT and the final answer) achieves 37.6\% accuracy at the cost of efficiency, producing an average response length of 8,740 tokens. In an attempt to elicit responses with shorter reasoning, this model could not provide short answers when instructed not to think (``No-Think'' mode), only reducing the  average response length to 5,134 tokens.

\textbf{System-1 Distillation\ } When training with System-1 demonstrations from the teacher model (i.e. only the final answers without reasoning), the student model achieves significant inference-time efficiency gains. We observe that responses are 27x shorter, at the cost of only a 4.6\% drop in accuracy compared to System-2 Distillation. 
However, we find that these models lack the ability to leverage test-time compute, even when explicitly instructed to spend more token budget for reasoning in the Adaptive-Think mode.

\textbf{Random Mixing Distillation\ }
Training with a random mix of System-1 and System-2 reasoning enables the student model to achieve flexible test-time compute by interpolating between thinking fast and slow. For example, when training with 40\% System-2 responses ($P_{System2}=0.4$), the average response length drops from 9,155 tokens in the ``Think'' mode to 1,642 and 847 for the ``Adaptive-Think'' and ``No-Think'' modes, respectively. Increasing the ratio of System-2 examples improves performance but increases response lengths. For example, models trained with 60\% System-2 achieve an accuracy of 37.5\% in the ``Think'' mode, compared to 36.7\% and 34.7\% for training with 40\% and 20\% System-2 reasoning respectively.

\textbf{Difficulty Based Mixing Distillation\ }
In contrast to random mixing, the difficulty-based mixing approach selectively distills System-2 responses for difficult questions and System-1 responses for easy questions. 
This targeted strategy, like random mixing, enables the model to flexibly adapt its response length at inference time.
Notably, the difficulty-based mixing method achieves an accuracy of 38.9\%, representing a 1.3\% improvement over System-2 distillation.
Furthermore, this approach results in only 36\% System-2 responses in the training set, yet it surpasses all random mixing methods, including the one with 60\% System-2 responses, which yields an accuracy of 37.5\%. This suggests that by selectively applying System-1 and System-2 distillation based on question difficulty, the model can strike a better balance between accuracy and efficiency.

%% file: meta_template_arxiv/sections/conclusion.tex
\section{Ablations and Analysis }

\begin{wrapfigure}{r}{0.45\textwidth}
\vspace{-2.5em}
    \centering
    \subfloat{{\includegraphics[width=0.95\linewidth]{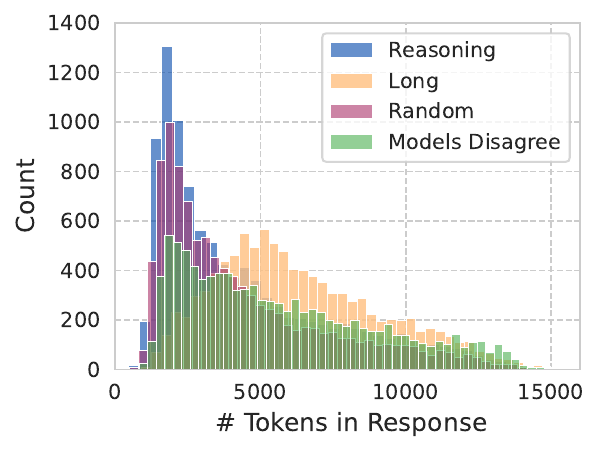} }}%
    \vspace{-1em}
    \caption{Length distributions of data subsets obtained from different selection methods.}
    \label{fig:length_distribution}
\vspace{-2em}
\end{wrapfigure}

\textbf{Long is More?\ } Prior work on data selection for LLM alignment has shown that selecting long samples is a tough-to-beat baseline ~\citep{zhao2024longalignmentsimpletoughtobeat}. Through our experiments, we validate that length is also a tough-to-beat baseline for LLM reasoning. However, as is shown in \autoref{fig:length_distribution}, we find that the most performant selection strategies (Long, Reasoning Strategies and Models Disagree) all have different distributions of reasoning length, indicating that length is not the only causal factor. Furthermore, selection based on reasoning strategy diversity yields a very similar length distribution to that of random selection, yet the former leads to better performance.

\textbf{Reducing Sequence Length: Improvements with Mixed Distillation\ }
One natural question is whether training with shorter reasoning traces achieves the same efficiency gains as mixed distillation. To this end, we experiment with reducing the maximum sequence length of training examples from 16,384 to 8,192 tokens, and use the same evaluation setting as described in \autoref{sec:distillation_methods}. Results are provided in \autoref{tab:mixed_think_8k}.

As expected, this leads to shorter responses at inference time, with an average length of 440 tokens in the Adaptive-Think mode and 372 tokens in No-Think mode. However, it comes at the cost of a severe drop in accuracy, with System-2 distillation scoring 32.5\% and 32.1\% in those two modes respectively. In contrast, applying difficulty-based mixing with a sequence length of 8,192 tokens not only preserves the ability to adjust response length via Adaptive Thinking but also yields notable accuracy improvements. Specifically, we observe gains of 0.6\% over System-2 distillation with a sequence length of 16,384 and 1.2\% over System-2 distillation with a sequence length of 8,192, while producing 2x fewer tokens. This suggests that difficulty-based mixing can effectively balance efficiency and accuracy, even at reduced sequence lengths.

\textbf{Effect of easy-to-verify data} We further evaluate the impact of training on problems with easy-to-verify answers. We use selection based on "Models Disagree" as a baseline, to compare with additionally selecting problems whose reference answers have short lengths ($\leq 9$ words) as a proxy for "easy-to-verify". As is shown in \autoref{tab:ref_ans}, training on reasoning traces of such problems does not bring consistent improvements except on math reasoning tasks.
\begin{table*}[h]
    \centering
    \small
    \renewcommand{\arraystretch}{0.9} %
    
    \begin{adjustbox}{max width=0.95\linewidth}
        \begin{tabular}{lccccc}
            \toprule
            & Train Size &  GPQA-D & MATH-500 & MMLU Pro & SuperGPQA \\

            \midrule
            Models Disagree & 10k& $39.7_{\pm0.4}$ & $61.9_{\pm0.3}$ & 59.1 & 29.7\\
            \hspace{3mm}with short reference answer  & 10k & $39.0_{\pm0.7}$ & $63.0_{\pm0.3}$ & 59.2 & 28.1 \\
            Models Disagree & 100k & $45.2_{\pm0.6}$ & $70.2_{\pm0.2}$ & 59.8 & 32.2 \\
            \hspace{3mm}with short reference answer & 100k & $43.8_{\pm0.5}$ & $71.4_{\pm0.4}$ & 61.4 & 31.2 \\
            \bottomrule
        \end{tabular}
    \end{adjustbox}
    \caption{\textbf{Reference-answer-based selection.} We show that filtering questions based on the length of the reference answers (as a proxy for easy-to-verify problems), in addition to disagreement selection, yields improvement in math reasoning performance.}
    \vspace{0.2cm}
    \label{tab:ref_ans}
    \vspace{-0.10cm}
\end{table*}

\section{Discussion and Conclusion}

In this work, we conduct an in-depth analysis of aspects of training data that lead to better scaling performance and sample efficiency when distilling reasoning capabilities from a strong teacher to a weaker student model. Our results offer new insights on the selection and curation of reasoning data, where current understanding mostly focuses on the overall importance of data quality, e.g. the ``Less is More'' hypothesis. Our experiments indicate that for ``learning to reason'', the diversity of reasoning primitives matters more than the diversity of topics or domains. Questions that are more difficult usually elicit more reasoning steps and thus serve as better demonstration examples to distill reasoning skills. We attribute lagging improvements on the mathematical task evaluations to the fact that the majority of the problems in our source prompt data do not look like problems appearing in those math evaluations.

Insights derived from our experiments and analysis are highly relevant to the growing interest in building small reasoning models. The training recipe of state-of-the-art small reasoning models \citep{yang2025qwen3technicalreport} also emphasizes the importance of data curation used in both RL post-training and strong-to-weak SFT distillation~\citep{bercovich2025llamanemotronefficientreasoningmodels}. In practice, a few criteria are used to select samples that (i) are as challenging as possible and (ii) cover a broad range of sub-domains. Our findings are complementary to those, where we demonstrate the potential performance gains from having explicit control over which fine-grained reasoning primitives to distill from the teacher model. Furthermore, we demonstrate steerability of reasoning efficiency in our mixed System-1/System-2 distillation experiments.

\textbf{Limitations\ } Our experiments are conducted in the setting of off-policy distillation where the student model is trained on the cross entropy loss from the teacher's labels. Another commonly adopted approach in knowledge distillation is on-policy distillation \citep{agarwal2023onpolicy}, where the student model is trained to match the logits of the teacher model computed given a partial context generated by the student. Future work should verify whether the same findings hold in the on-policy  setting. The implications of selecting distillation examples can be further explored by conducting reinforcement learning following the SFT distillation stage with curated reasoning data.

%% file: meta_template_arxiv/sections/appendix.tex
\newpage
\clearpage

\appendix \section{Prompts used in Annotation}
\label{app:prompts}
The prompt used for annotating reasoning traces to obtain reasoning strategies and verbosity score is provided in \autoref{tab:prompt_annotate_reasoning}.

\autoref{tab:prompt_annotate_domain_5} covers the prompt used to annotate disciplines and fields for questions in the dataset.

\begin{figure}[ht]
    \centering
    \includegraphics[width=0.6\linewidth]{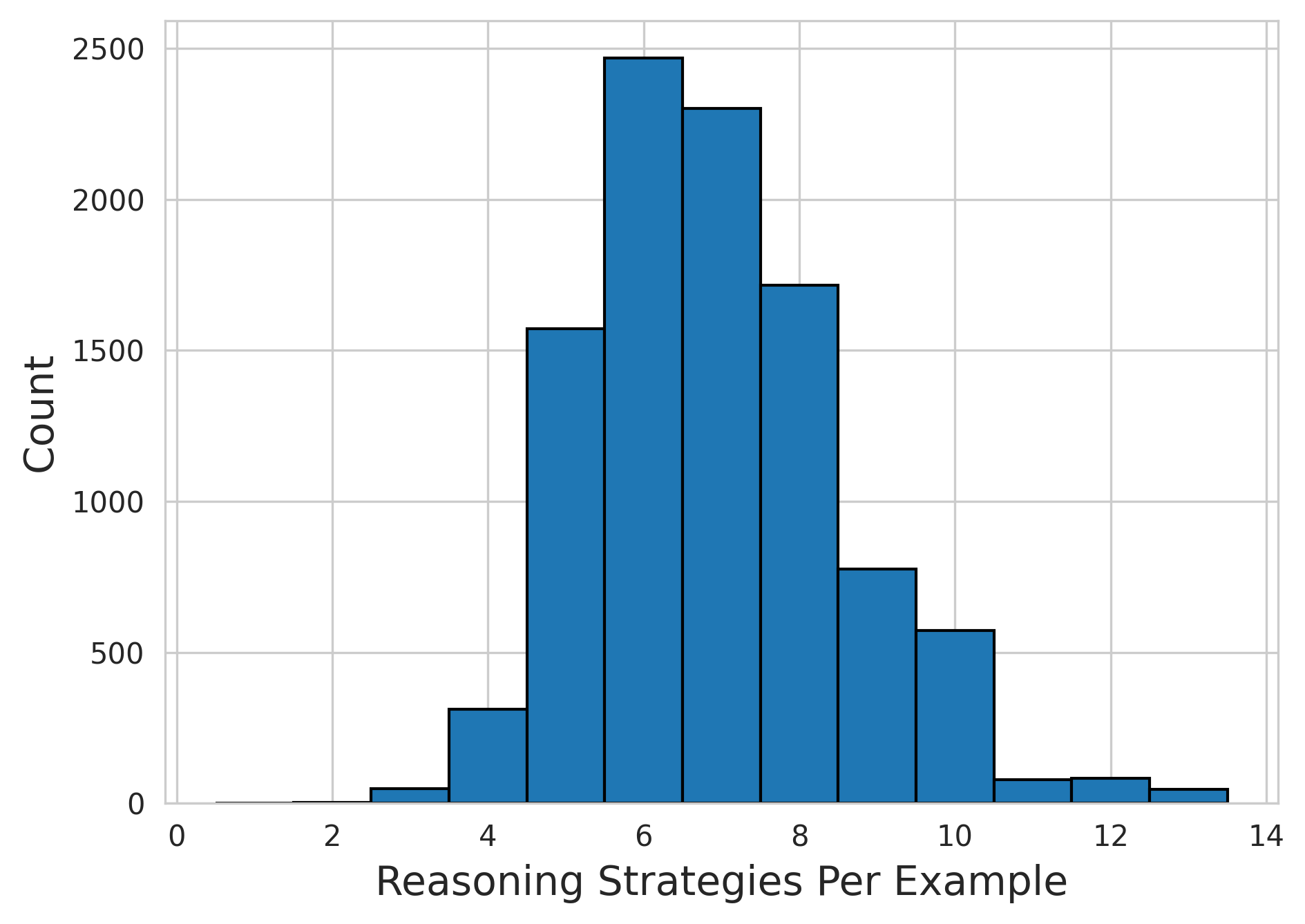}
    \vspace{-0.2cm}
    \caption{Distribution of unique reasoning strategies per example based on annotations of 10,000 samples.}
    \label{fig:unique_reasoning_hist}
\end{figure}

\begin{figure*}[th!]
\centering
\small
\begin{tcolorbox}[colback=green!10!white, %
                  colframe=green!30!white, %
                  width=0.99\linewidth, %
                  arc=4mm, %
                  auto outer arc,
                  before upper={\setlength{\parindent}{0em}} %
                  ]

\begin{lstlisting}[basicstyle=\ttfamily, breaklines=true, showstringspaces=false, breakindent=0pt]
Below is a question and solution generated by an LLM. Your task is to summarize the problem-solving steps used by the LLM. Read the thought process carefully, and annotate the explorations in the thought process used by the LLM. Specifically, write down detailed steps the LLM took to pursue its thinking process, identifying all meta-reasoning strategies used at each step, e.g. self-verification, backtracking, exploration, etc. Based on these analysis, also check the degrees of verbosity of the reasoning traces, e.g. how much unnecessary ramblings were found during the thinking process which does not make much progress in solving the problem. Derive a verbosity_score in the end. The verbosity_score should be derived on a scale of 0 to 10. Score 0 means the problem solving in the thinking process is very efficient with no rambling at all. Score 10 means the reasoning traces are very verbose, where the thinking process is long but each step does not make progress in solving the problem. Organize your answer in a json so that the steps and meta-reasoning strategies and the final verbosity_score can be easily extracted.
Question: {question}
Solution from LLM: {reasoning trace to be annotated}}
\end{lstlisting}
\end{tcolorbox}

\caption{Prompt for annotating reasoning traces to obtain reasoning strategies and the verbosity score.}
\label{tab:prompt_annotate_reasoning}
\end{figure*}

\begin{figure*}[htbp!]
\centering
\small
\begin{tcolorbox}[colback=green!10!white, %
                  colframe=green!30!white, %
                  width=0.99\linewidth, %
                  arc=4mm, %
                  auto outer arc,
                  sharp corners=south, %
                  before upper={\setlength{\parindent}{0em}} %
                  ]
\begin{lstlisting}[basicstyle=\ttfamily, breaklines=true, showstringspaces=false, breakindent=0pt]
You are an expert in labeling questions into categories. 

For a given question, use the following taxonomy for labelling, which is structured as {'discipline': {'field': ['sub-field', 'sub-field', ...]}}. 

{'Engineering': {'Electronic Science and Technology': ['Circuits and Systems',
   'Microelectronics and Solid-State Electronics',
   'Electromagnetic Field and Microwave Technology'],
  'Computer Science and Technology': ['Computer Architecture',
   'Data Structures',
   'Operating Systems',
   'Computer Software and Theory',
   'Advanced Programming Languages',
   'Pattern Recognition',
   'Principles of Computer Organization',
   'Computer Networks',
   'Databases',
   'Formal Languages'],
  'Information and Communication Engineering': ['Signal and Information Processing',
   'Optical Fiber Communication',
   'Communication and Information Systems',
   'Antenna and Radio Communication',
   'Communication Principles'],
  'Control Science and Engineering': ['Control Theory and Control Engineering',
   'Operations Research and Cybernetics',
   'Guidance, Navigation and Control'],
  'Materials Science and Engineering': ['Materials Physics and Chemistry',
   'Materials Processing Engineering'],
  'Electrical Engineering': ['Power Electronics and Electrical Drives',
   'Electrical Theory and New Technologies',
   'High Voltage and Insulation Technology',
   'Power Systems and Automation'],
  'Power Engineering and Engineering Thermophysics': ['Power Machinery and Engineering',
   'Internal Combustion Engineering',
   'Thermal Energy Engineering',
   'Engineering Thermophysics',
   'Refrigeration and Cryogenic Engineering',
   'Fluid Machinery and Engineering',
   'Heat Transfer',
   'Engineering Fluid Mechanics'],
  'Hydraulic Engineering': ['Water conservancy and Hydropower Engineering',
   'Hydraulics and Hydrology'],
  'Chemical Engineering and Technology': ['Mass Transport and Separation Process in Chemical Engineering',
   'Fluid Flow and Heat Transfer in Chemical Engineering',
   'Elements of Chemical Reaction Engineering',
   'Chemical Transport Engineering'],
  'Architecture': ['Urban Planning and Design',
   'Architectural Design and Theory',
   'Architectural History'],
  'Forestry Engineering': ['Forest Engineering',
   'Wood Science and Technology'],
\end{lstlisting}
\end{tcolorbox}
\captionsetup{labelformat=empty}
\label{tab:prompt_annotate_domain_1}
\end{figure*}
\hfill %

\begin{figure*}[htbp!]
\centering
\small
\begin{tcolorbox}[colback=green!10!white, %
                  colframe=green!30!white, %
                  width=0.99\linewidth, %
                  arc=4mm, %
                  auto outer arc,
                  sharp corners=north, 
                  sharp corners=south,
                  before upper={\setlength{\parindent}{0em}} %
                  ]
\begin{lstlisting}[basicstyle=\ttfamily, breaklines=true]
  'Nuclear Science and Technology': ['Radiation Protection and Nuclear Technology Applications',
   'Nuclear Energy and Reactor Technology'],
  'Weapon Science and Technology': ['Weapon Systems Science and Engineering',
   'Military Chemistry and Pyrotechnics'],
  'Naval Architecture and Ocean Engineering': ['Marine Engineering',
   'Ship Mechanics and Design Principles'],
  'Environmental Science and Engineering': ['Environmental and Resource Protection',
   'Environmental Engineering',
   'Environmental Science'],
  'Transportation Engineering': ['Vehicle Operation Engineering',
   'Traffic Information Engineering and Control',
   'Transportation Planning and Management',
   'Road and Railway Engineering'],
  'Mechanical Engineering': ['Manufacturing Automation',
   'Mechatronic Engineering'],
  'Aeronautical and Astronautical Science and Technology': ['Aeronautical and Astronautical Science and Technology'],
  'Civil Engineering': ['Geotechnical Engineering',
   'Structural Engineering',
   'Bridge and Tunnel Engineering',
   'Urban Infrastructure Engineering'],
  'Mechanics': ['Fundamentals of Dynamics and Control',
   'Theoretical Fluid Mechanics',
   'Theoretical Mechanics',
   'Rigid Body Mechanics',
   'Solid Mechanics'],
  'Petroleum and Natural Gas Engineering': ['Poromechanics and Reservoir Physics',
   'Oil and Gas Field Development and Storage & Transportation Engineering'],
  'Food Science and Engineering': ['Food Processing and Storage Engineering',
   'Food Biochemistry'],
  'Agricultural Engineering': ['Agricultural Environment and Soil-Water Engineering',
   'Agricultural Mechanization Engineering'],
  'Surveying and Mapping Science and Technology': ['Geodesy and Surveying Engineering',
   'Cartography and Geographic Information Engineering',
   'Digital Surveying and Remote Sensing Applications'],
  'Metallurgical Engineering': ['Iron and Steel Metallurgy',
   'Principles of Metallurgy',
   'Non-ferrous Metallurgy',
   'Physical Chemistry of Metallurgical Process'],
  'Mining Engineering': ['Mining and Safety Engineering',
   'Mineral Processing Engineering'],
  'Geological Resources and Geological Engineering': ['Geological Resources and Geological Engineering'],
  'Optical Engineering': ['Theoretical Optics',
   'Optoelectronic Technology',
   'Laser Technology',
   'Applied Optics'],
  'Textile Science and Engineering': ['Textile Materials Science',
   'Textile Chemistry and Dyeing Engineering'],
  'Instrument Science and Technology': ['Instrument Science and Technology']},
\end{lstlisting}
\end{tcolorbox}
\captionsetup{labelformat=empty}
\label{tab:prompt_annotate_domain_2}
\end{figure*}
\hfill %

\begin{figure*}[htbp!]
\centering
\small
\begin{tcolorbox}[colback=green!10!white, %
                  colframe=green!30!white, %
                  width=0.99\linewidth, %
                  arc=4mm, %
                  auto outer arc,
                  sharp corners=north, 
                  sharp corners=south, 
                  before upper={\setlength{\parindent}{0em}} %
                  ]
\begin{lstlisting}[basicstyle=\ttfamily, breaklines=true]
'Philosophy': {'Philosophy': ['Philosophical Aesthetics',
   'Ethics',
   'Logic',
   'Philosophy of Science and Technology',
   'Religious Studies']},
'Medicine': {'Traditional Chinese Medicine': ['Traditional Chinese Medicine Theory',
   'Traditional Chinese Health Preservation',
   'Traditional Chinese Pharmacy'],
  'Clinical Medicine': ['Internal Medicine',
   'Obstetrics and Gynecology',
   'Emergency Medicine',
   'Neurology',
   'Psychiatry and Mental Health',
   'Surgery',
   'Imaging and Nuclear Medicine',
   'Otorhinolaryngology',
   'Dermatology and Venereology',
   'Ophthalmology',
   'Geriatric Medicine',
   'Oncology',
   'Clinical Laboratory Diagnostics',
   'Anesthesiology',
   'Pediatrics',
   'Nursing and Rehabilitation Medicine'],
  'Basic Medicine': ['Pathogen Biology',
   'Immunology',
   'Human Anatomy and Histology-Embryology',
   'Pathology and Pathophysiology',
   'Forensic Medicine',
   'Radiation Medicine'],
  'Public Health and Preventive Medicine': ['Epidemiology and Health Statistics',
   'Health Toxicology and Environmental Health',
   'Maternal, Child and Adolescent Health',
   'Nutrition and Food Hygiene'],
  'Pharmacy': ['Pharmacology',
   'Microbiology and Biochemical Pharmacy',
   'Pharmaceutical Analysis',
   'Medicinal Chemistry',
   'Pharmaceutics'],
  'Stomatology': ['Basic Stomatology', 'Clinical Stomatology']},
'Economics': {'Applied Economics': ['Finance',
   'Public Finance',
   'International Trade',
   'Labor Economics',
   'Economic Statistics',
   'Quantitative Economics',
   'Industrial Economics',
   'National and Defense Economics'],
  'Theoretical Economics': ['Political Economy',
   'Economic History',
   'Western Economics']},
  
\end{lstlisting}
\end{tcolorbox}
\captionsetup{labelformat=empty}
\label{tab:prompt_annotate_domain_3}
\end{figure*}
\hfill %

\begin{figure*}[htbp!]
\centering
\small
\begin{tcolorbox}[colback=green!10!white, %
                  colframe=green!30!white, %
                  width=0.99\linewidth, %
                  arc=4mm, %
                  auto outer arc,
                  sharp corners=north, 
                  sharp corners=south, 
                  before upper={\setlength{\parindent}{0em}} %
                  ]
\begin{lstlisting}[basicstyle=\ttfamily, breaklines=true]
'Science': {'Mathematics': ['Combinatorial Mathematics',
   'Ordinary Differential Equations',
   'Mathematical Analysis',
   'Advanced Algebra',
   'Functions of Real Variables',
   'Probability and Statistics',
   'Numerical Analysis',
   'Polynomials and Series Expansions',
   'Geometry and Topology',
   'Computational Mathematics',
   'Discrete Mathematics',
   'Stochastic Processes',
   'Functions of Complex Variables',
   'Fundamental Mathematics',
   'Number Theory',
   'Group Theory',
   'Cryptography',
   'Fuzzy Mathematics',
   'Special Number Theory',
   'Graph Theory'],
  'Physics': ['Particle and Nuclear Physics',
   'Electrodynamics',
   'Quantum Mechanics',
   'Atomic and Molecular Physics',
   'Fluid Physics',
   'Solid State Physics',
   'Relativity',
   'Thermodynamics and Statistical Physics',
   'Subatomic and Atomic Physics',
   'Semiconductor Physics',
   'Polymer Physics',
   'Statistical Mechanics',
   'Thermodynamics',
   'Acoustics'],
  'Systems Science': ['Systems Science'],
  'Biology': ['Botany',
   'Biochemistry and Molecular Biology',
   'Genetics',
   'Zoology',
   'Biophysics',
   'Cell Biology',
   'Physiology',
   'Microbiology',
   'Ecology'],
  'Chemistry': ['Organic Chemistry',
   'Physical Chemistry',
   'Analytical Chemistry',
   'Electrochemistry',
   'Radiochemistry',
   'Polymer Chemistry and Physics',
   'Inorganic Chemistry'],
  'Geography': ['Human Geography', 'Physical Geography'],
  'Oceanography': ['Hydrogeology',
   'Marine Chemistry',
   'Underwater Acoustics',
   'Marine Biology'],
\end{lstlisting}
\end{tcolorbox}
\captionsetup{labelformat=empty}
\label{tab:prompt_annotate_domain}
\end{figure*}
\hfill %

\begin{figure*}[htbp!]
\centering
\small
\begin{tcolorbox}[colback=green!10!white, %
                  colframe=green!30!white, %
                  width=0.99\linewidth, %
                  arc=4mm, %
                  auto outer arc,
                  sharp corners=north, 
                  sharp corners=south, 
                  before upper={\setlength{\parindent}{0em}} %
                  ]
\begin{lstlisting}[basicstyle=\ttfamily, breaklines=true]
  'Geology': ['Principles of Seismic Exploration',
   'Structural Geology',
   'Mineralogy, Petrology, and Economic Geology',
   'Paleontology and Stratigraphy',
   'Geochemistry'],
  'Physical Oceanography': ['Physical Oceanography'],
  'Astronomy': ['Solar System Science',
   'Astrophysics',
   'Stellar and Interstellar Evolution',
   'Astronomical Observation and Technology',
   'Cosmology'],
  'Atmospheric Science': ['Atmospheric Physics and Atmospheric Environment',
   'Meteorology',
   'Dynamic Meteorology'],
  'Geophysics': ['Solid Earth Geophysics', 'Space physics']},
'Law': {'Law': ['Procedural Law',
   'Constitutional and Administrative Law',
   'Criminal Law',
   'Civil and Commercial Law',
   'Contract Law',
   'Military Law',
   'Law and Social Governance',
   'International Law',
   'Legal Theory and Legal History'],
  'Political Science': ['Political Science']},
'History': {'History': ['World History',
   'Historical Geography',
   'Archaeology and Museology']},
'Education': {'Education': ['Theory of Curriculum and Instruction',
   'Educational Technology and Principles',
   'Special Education',
   'Preschool Education'],
  'Psychology': ['Psychology'],
  'Physical Education': ['Sports Science and Medicine',
   'Sports Humanities and Sociology',
   'Physical Education and Training']},
'Military Science': {'Military Science': ['Military Thought and History',
   'Military Command and Information Systems',
   'Military Management',
   'Military Logistics and Equipment']},
'Management': {'Business Administration': ['Business and Accounting Management',
   'Tourism Management and Technological Economics Management'],
  'Public Administration': ['Social Medicine and Health Management',
   'Land Resource Management and Administrative Management',
   'Education Economics, Management and Social Security'],
  'Library, Information and Archival Management': ['Information Management and Communication',
   'Library and Archival Science',
   'Information Management Science'],
  'Management Science and Engineering': ['Management Science and Engineering']},
\end{lstlisting}
\end{tcolorbox}
\captionsetup{labelformat=empty}
\label{tab:prompt_annotate_domain_4}
\end{figure*}
\hfill %

\begin{figure*}[htbp!]
\centering
\small
\begin{tcolorbox}[colback=green!10!white, %
                  colframe=green!30!white, %
                  width=0.99\linewidth, %
                  arc=4mm, %
                  auto outer arc,
                  sharp corners=north, 
                  before upper={\setlength{\parindent}{0em}} %
                  ]
\begin{lstlisting}[basicstyle=\ttfamily, breaklines=true, showstringspaces=false, breakindent=0pt]
  'Literature and Arts': {'Language and Literature': ['Literary History',
   'French Language and Literature',
   'Linguistics and Applied Linguistics',
   'Literary Theory',
   'Philology and Bibliography',
   'Modern and Contemporary Chinese Literature',
   'Classical Chinese Literature',
   'Russian Language and Literature'],
  'Art Studies': ['Dance Studies',
   'Design Arts',
   'Broadcasting and Television Art',
   'Fine Arts',
   'Drama and Opera Studies',
   'Film Studies'],
  'Journalism and Communication': ['Journalism and News Practice',
   'History and Theory of Journalism and Media Management',
   'Communication and Broadcasting'],
  'Musicology': ['Composition',
   'Instrumentation and Performance',
   'Music History, Education, and Technology',
   'Musical Forms and Analysis',
   'Harmony',
   'Pitch and Scales']},
'Agronomy': {'Aquaculture': ['Aquaculture'],
  'Animal Husbandry': ['Animal Rearing and Breeding',
   'Animal Nutrition and Feed Science'],
  'Crop Science': ['Crop Science'],
  'Forestry': ['Forest Cultivation and Genetic Breeding',
   'Landscape Plants and Ornamental Horticulture'],
  'Veterinary Medicine': ['Veterinary Medicine']},
'Sociology': {'Sociology': ['Social and Folklore Studies',
   'Demography and Anthropology']}}
First, identify the most appropriate discipline. Then identify the most appropriate filed within the chosen discipline. Finally, identify the most appropriate sub-filed.

Put your final labelling results in a json object:
```
{
   "discipline": <choose the most appropriate discipline>, 
   "field": <choose the most appropriate field within the discipline>, 
    "sub_field": <choose the most appropriate sub_field within the field>
}
```

Under no circumstances use enumeration and do not give more than 1 discipline, 1 field and 1 sub-field and do not use any labels that are not in above dictionary. 

### Question

\end{lstlisting}
\end{tcolorbox}
\caption{Prompt for annotating question with discipline and fields.}
\label{tab:prompt_annotate_domain_5}
\end{figure*}

\newpage

\section{Diversity using Semantic Embeddings based Clustering}
\label{app:clustering}

In \autoref{tab:embeddings}, we compare the performance of different clustering methods. Three popular embedding models are considered: all-MiniLM-L6-v2~\citep{wang2020minilmdeepselfattentiondistillation}, SimCSE~\citep{gao2022simcsesimplecontrastivelearning} and Llama-3.1-8B-Instruct~\citep{grattafiori2024llama3herdmodels}. The embedding size of all-MiniLM-L6-v2 is 384, sup-simcse-roberta-large is 1024 and Llama-3.1-8B-Instruct is 4096. After choosing the best embedding model Llama-3.1-8B, we then experiment with different clustering methods. We use K-means with different numbers of clusters: 100, 500, 1k, 10k. We also experiment with density-based HDBSCAN~\citep{mcinnes2017hdbscan} clustering method. In each experiment, we use equal stratified sampling to randomly sample same number of items from each cluster. The best combination of Llama-3.1-8B-Instruct with HDBSCAN is reported in the main experiment \autoref{tab:selection-8b}.

\begin{table*}[!t]
    \centering
    \renewcommand{\arraystretch}{0.9} %
    \caption{Comparison of different semantic embedding clustering methods for 10k training examples.}
    \begin{adjustbox}{max width=1\linewidth}
        \begin{tabular}{lccccccr}
            \toprule
            Embeddings & Clustering & \# clusters & GPQA-D & MATH-500 & MMLU Pro & SuperGPQA & Average \\
            \midrule
            all-MiniLM-L6-v2 & K-means & 10k & $36.3_{\pm0.8}$ & $57.5_{\pm0.4}$ & 55.4 & 25.9 & 43.8 \\
            sup-simcse-roberta-large & K-means & 10k & $34.3_{\pm0.7}$ & $58.8_{\pm0.4}$ & 56.1 & 26.7 & 44.0 \\
            \midrule
            \multirow{4}{*}{Llama-3.1-8B-Instruct} & \multirow{3}{*}{K-means} & 100 & $39.4_{\pm0.7}$ & $57.6_{\pm0.4}$ & 55.5 & 26.1 & 44.7 \\
            & & 500 & $36.0_{\pm0.5}$ & $59.6_{\pm0.3}$ & 56.4 & 26.5 & 44.6\\
            & & 1k & $35.8_{\pm0.8}$ & $59.8_{\pm0.3}$ & 56.5 & 26.7 & 44.7 \\
            & & 10k & $36.5_{\pm0.6}$ & $59.0_{\pm0.3}$ & 57.3 & 26.2 & 44.8 \\
            \cmidrule(lr){2-8}
            & HDBSCAN & 991 & $39.4_{\pm0.5}$ & $60.3_{\pm0.3}$ & 55.9 & 27.2 & 45.7 \\
            \bottomrule
        \end{tabular}
    \end{adjustbox}
    \label{tab:embeddings}
\end{table*}

\begin{table*}[!t]
    \centering
    \renewcommand{\arraystretch}{0.9} %
        \caption{Comparison of different methods for mixing System-1 and System-2 responses using sequence lengths of 8K and 16K.}
        \vspace{-0.2cm}
      \begin{adjustbox}{max width=1.0\textwidth}
        \begin{tabular}{lcccccccc}
            \toprule

            &
            &\multicolumn{2}{c}{\textbf{No-Think}} &\multicolumn{2}{c}{\textbf{Adaptive-Think}} &\multicolumn{2}{c}{\textbf{Think}} 
            \\
            \cmidrule(lr){3-4} \cmidrule(lr){5-6} \cmidrule(lr){7-8}
            & Sequence Length  & Mean Length  & Accuracy & Mean Length  & Accuracy &  Mean Length  & Accuracy  \\\\
            \midrule
            System-1  & 8192 &250.8  & $34_{\pm0.5}$ & 533.4 & $33.1_{\pm0.5}$ & - & -\\
            System-1   & 16384 & 321.3 & $34_{\pm0.6}$ & 541.1 & $32.4_{\pm0.7}$ & - & -\\
            \midrule
            System-2  & 8192 & 371.6 & $32.1_{\pm0.5}$ & 439.5 & $32.5_{\pm0.5}$ & 8510.2 &$37_{\pm0.6}$ \\
            System-2   & 16384 & 5133.7 & $36.3_{\pm0.5}$ & 7817 & $37.3_{\pm0.6}$ & 8740.6 &$37.6_{\pm0.6}$ \\

            \midrule
            Difficulty-based Mixing & 8192 &243.6 &$33.5_{\pm0.6}$  &904.6 &$35.15_{\pm0.5}$ &4735.4 & $38.2_{\pm0.5}$ \\
            Difficulty-based Mixing & 16384 & 799.4 &$34.5_{\pm0.5}$  &2033 &$35.1_{\pm0.6}$ &7562.4 & $38.9_{\pm0.7}$ \\
            \bottomrule
        \end{tabular}
        \end{adjustbox}
        \label{tab:mixed_think_8k}
        \vspace{-0.10cm}
\end{table*}

\section{Reasoning strategy count ablation}
\label{app:reasoning_strategies}
The number of unique reasoning strategies found per training example is shown in \autoref{fig:unique_reasoning_hist}. We split training examples into bucket Low (less than 5 strategies), Med (5-8 strategies) and High (more than 8 strategies) and perform SFT distillation. From \autoref{tab:strategy_count} we observe that on average the Med category performs the best. The fact that performance varies on different buckets indicates that the reasoning strategy should align with specific tasks.

\begin{table*}[ht]
    \centering
    \caption{Comparison of different reasoning strategy count sampling methods.}
    \vspace{-0.2cm}
    \begin{adjustbox}{max width=0.95\linewidth}
        \begin{tabular}{lccccccr}
            \toprule
            sampling method & \# examples & GPQA-D & MATH-500 & MMLU Pro & SuperGPQA \\
            \midrule
            \# Strategies=Low & 10k& $34.9_{\pm0.4}$ & $60.9_{\pm0.2}$ & 57.2 & 27.5 \\
            \# Strategies=Med & 10k& $40.3_{\pm0.6}$ & $59.9_{\pm0.3}$ & 56.8 & 28.2\\
            \# Strategies=High & 10k& $38.2_{\pm0.5}$ & $60.8_{\pm0.3}$ & 57.4 & 27.6\\
            \bottomrule
        \end{tabular}
    \end{adjustbox}
    \label{tab:strategy_count}
    \vspace{-0.10cm}
\end{table*}

\hfill %

\begin{figure*}[htbp!]
\centering
\small
\begin{tcolorbox}[colback=green!10!white, %
                  colframe=green!30!white, %
                  width=0.99\linewidth, %
                  arc=4mm, %
                  auto outer arc,
                  sharp corners=south, %
                  before upper={\setlength{\parindent}{0em}} %
                  ]
\begin{lstlisting}[basicstyle=\ttfamily, breaklines=true]
{'Mathematics': (1338522, 49.92),
 'Physics': (443010, 16.52),
 'Computer Science and Technology': (196323, 7.32),
 'Applied Economics': (144020, 5.37),
 'Biology': (60535, 2.26),
 'Chemistry': (51152, 1.91),
 'Law': (45345, 1.69),
 'Electronic Science and Technology': (43368, 1.62),
 'Education': (42881, 1.6),
 'Philosophy': (38093, 1.42),
 'Mechanical Engineering': (34756, 1.3),
 'Electrical Engineering': (24703, 0.92),
 'Clinical Medicine': (16956, 0.63),
 'Basic Medicine': (16256, 0.61),
 'Astronomy': (15106, 0.56),
 'Power Engineering and Engineering Thermophysics': (14027, 0.52),
 'Mechanics': (12406, 0.46),
 'Business Administration': (11542, 0.43),
 'Theoretical Economics': (10922, 0.41),
 'Information and Communication Engineering': (9936, 0.37),
 'Aeronautical and Astronautical Science and Technology': (8535, 0.32),
 'Civil Engineering': (7997, 0.3),
 'Control Science and Engineering': (7701, 0.29),
 'Language and Literature': (7255, 0.27),
 'Public Health and Preventive Medicine': (6363, 0.24),
 'Management Science and Engineering': (6072, 0.23),
 'Psychology': (5744, 0.21),
 'History': (5271, 0.2),
 'Geography': (4998, 0.19),
 'Sociology': (4482, 0.17),
 'Materials Science and Engineering': (4458, 0.17),
 'Optical Engineering': (3987, 0.15),
 'Chemical Engineering and Technology': (3541, 0.13),
 'Geology': (3340, 0.12),
 'Hydraulic Engineering': (3069, 0.11),
 'Transportation Engineering': (2621, 0.1),
 'Musicology': (2178, 0.08),
 'Crop Science': (2176, 0.08),
 'Art Studies': (1966, 0.07),
 'Atmospheric Science': (1955, 0.07),
 'Physical Education': (1939, 0.07),
 'Nuclear Science and Technology': (1836, 0.07),
 'Pharmacy': (1572, 0.06),
 'Geophysics': (1493, 0.06),
 'Environmental Science and Engineering': (1269, 0.05),
 'Systems Science': (1106, 0.04),
 'Political Science': (804, 0.03),
 'Naval Architecture and Ocean Engineering': (727, 0.03),
 'Animal Husbandry': (706, 0.03),
 'Public Administration': (651, 0.02),
 'Military Science': (593, 0.02),
 'Oceanography': (576, 0.02),
 'Architecture': (532, 0.02),
 'Food Science and Engineering': (482, 0.02),
 'Petroleum and Natural Gas Engineering': (441, 0.02),
 
\end{lstlisting}
\end{tcolorbox}
\captionsetup{labelformat=empty}
\label{tab:Domain_distribution_1}
\end{figure*}
\hfill %

\begin{figure*}[htbp!]
\centering
\small
\begin{tcolorbox}[colback=green!10!white, %
                  colframe=green!30!white, %
                  width=0.99\linewidth, %
                  arc=4mm, %
                  auto outer arc,
                  sharp corners=north, %
                  before upper={\setlength{\parindent}{0em}} %
                  ]
\begin{lstlisting}[basicstyle=\ttfamily, breaklines=true]
 'Surveying and Mapping Science and Technology': (405, 0.02),
 'Journalism and Communication': (374, 0.01),
 'Textile Science and Engineering': (328, 0.01),
 'Stomatology': (275, 0.01),
 'Metallurgical Engineering': (266, 0.01),
 'Forestry': (170, 0.01),
 'Weapon Science and Technology': (137, 0.01),
 'Veterinary Medicine': (133, 0.0),
 'Instrument Science and Technology': (125, 0.0),
 'Aquaculture': (125, 0.0),
 'Library, Information and Archival Management': (110, 0.0),
 'Physical Oceanography': (98, 0.0),
 'Traditional Chinese Medicine': (69, 0.0),
 'Mining Engineering': (68, 0.0),
 'Agricultural Engineering': (63, 0.0),
 'Forestry Engineering': (48, 0.0),
 'Geological Resources and Geological Engineering': (24, 0.0)}
\end{lstlisting}
\end{tcolorbox}
\caption{Domain distribution(count and percentage) among all the questions.}
\label{tab:Domain_distribution_2}
\end{figure*}
\hfill %

\begin{figure*}[th!]
\centering
\small
\begin{tcolorbox}[colback=green!10!white, %
                  colframe=green!30!white, %
                  width=0.99\linewidth, %
                  arc=4mm, %
                  auto outer arc,
                  sharp corners=south, %
                  before upper={\setlength{\parindent}{0em}} %
                  ]
\begin{lstlisting}[basicstyle=\ttfamily, breaklines=true, showstringspaces=false, breakindent=0pt]
Question: A weight is attached to a string and spun vertically in a gravity field. Describe the resultant force acting on the weight at the leftmost and rightmost positions of its circular motion, and explain why the resultant force does not always point towards the center of the circle.

"Okay, so I have this problem where a weight is attached to a string and spun vertically in a gravity field. I need to describe the resultant force acting on the weight at the leftmost and rightmost positions of its circular motion. Also, I have to explain why the resultant force doesn't always point towards the center of the circle. Hmm, let me think about this step by step.

First, I remember that when an object moves in a circular path, it experiences a centripetal force directed towards the center of the circle. This centripetal force is necessary to keep the object moving in a circular path instead of a straight line. In many cases, like when spinning a ball horizontally, the tension in the string provides the centripetal force. But in this case, it's a vertical circle, so gravity must play a role here.

At different points in the vertical circle, the forces acting on the weight will be different because gravity is always acting downward, whereas the tension in the string will vary depending on the position. Let me try to visualize this. When the weight is at the top of the circle, both tension and gravity are acting downward. At the bottom, tension acts upward and gravity downward. At the leftmost and rightmost positions, which are the sides of the circle, the string is horizontal, right? So gravity is still acting downward, and tension is along the string towards the center. Wait, but if the weight is at the leftmost point, the center is to the right, so tension is to the right, gravity is down. So the resultant force would be the combination of tension and gravity. Similarly, at the rightmost point, tension is towards the left, and gravity is still down.

But the question is about the leftmost and rightmost positions. Let me focus on those. So in these positions, the weight is moving either upward or downward? Wait, at the leftmost point, if the weight is moving in a vertical circle, the direction of motion is either up or down, depending on whether it's moving clockwise or counterclockwise. Let's assume it's moving in a vertical circle in such a way that when it's at the leftmost point, it's moving downward, and at the rightmost point, it's moving upward. Wait, that might not be right. Let me clarify.

In vertical circular motion, when you spin something around, at the top, the velocity is to the left or right, and at the bottom, it's the opposite. So actually, at the leftmost point, the velocity is downward, and at the rightmost point, it's upward? Hmm, maybe. Alternatively, if we're considering the standard vertical circle where you start from the bottom and go up, then at the top, the velocity is to the left, at the bottom to the right, so the leftmost and rightmost points would have velocity upward and downward? Wait, perhaps it's better to parametrize the motion.

\end{lstlisting}
\end{tcolorbox}
\captionsetup{labelformat=empty}
\label{tab:high_resonign_1}
\end{figure*}

\begin{figure*}[th!]
\centering
\small
\begin{tcolorbox}[colback=green!10!white, %
                  colframe=green!30!white, %
                  width=0.99\linewidth, %
                  arc=4mm, %
                  auto outer arc,
                  sharp corners=north, 
                  sharp corners=south, %
                  before upper={\setlength{\parindent}{0em}} %
                  ]
\begin{lstlisting}[basicstyle=\ttfamily, breaklines=true, showstringspaces=false, breakindent=0pt]
Alternatively, perhaps more straightforwardly, at any point in the circular motion, the direction of the velocity is tangential to the circle. So at the leftmost point, the velocity is straight down, and at the rightmost point, it's straight up. Wait, no. If the circle is vertical, the top of the circle is highest point, bottom is lowest. Then, leftmost and rightmost points are at the sides. So when the object is at the leftmost point, moving down from the top to the bottom, the velocity direction is downward. Wait, actually, at the leftmost point, the velocity is in the downward direction? Wait, no, in a circle, the direction of velocity is tangent to the circle. So at the leftmost point, if you imagine a circle, the tangent is vertical, pointing down on the left and up on the right. Wait, no. Wait, maybe I should draw this mentally.

Consider a circle on a vertical plane. The top point has velocity to the left or right, depending on the direction of rotation. Suppose the weight is moving clockwise. Then at the top, velocity is to the right; at the bottom, to the left; at the leftmost point, velocity is downward; at the rightmost point, velocity is upward. If it's counterclockwise, then at the top, velocity is to the left; bottom to the right; leftmost upward; rightmost downward. So depending on direction, the velocity direction at those side points is either up or down. Hmm.

But regardless, in terms of the forces acting on the weight at the leftmost and rightmost positions: At each of these points, the forces acting are tension in the string and gravity. The tension is directed towards the center of the circle, which at leftmost position is towards the right, and at the rightmost position is towards the left. Gravity is always acting downward, with magnitude mg. So at these two points, the tension provides the centripetal force component, but gravity acts perpendicular to the tension? Wait, at the leftmost and rightmost points, gravity is vertical and tension is horizontal. Therefore, the two forces are perpendicular to each other at these points. Therefore, the resultant force would be the vector sum of the horizontal tension and the vertical gravitational force. So in that case, the resultant force is not purely towards the center, because you have two components: one towards the center (tension) and one downward (gravity). So the resultant force would be the combination of these two, hence not pointing towards the center. But wait, isn't centripetal acceleration still required? Let me think.

Wait, in uniform circular motion, the net force must be centripetal, but in non-uniform circular motion, such as when there is tangential acceleration, the net force has both centripetal and tangential components. In this case, when you spin a weight vertically, the speed is not constant because gravity does work on the object, speeding it up as it comes down and slowing it down as it goes up. Therefore, the motion is not uniform, so there is a tangential component of acceleration, which means the net force has both a centripetal component (towards the center) and a tangential component (tangent to the circle).

\end{lstlisting}
\end{tcolorbox}
\captionsetup{labelformat=empty}
\label{tab:high_resonign_2}
\end{figure*}

\begin{figure*}[th!]
\centering
\small
\begin{tcolorbox}[colback=green!10!white, %
                  colframe=green!30!white, %
                  width=0.99\linewidth, %
                  arc=4mm, %
                  auto outer arc,
                  sharp corners=north, 
                  sharp corners=south, %
                  before upper={\setlength{\parindent}{0em}} %
                  ]
\begin{lstlisting}[basicstyle=\ttfamily, breaklines=true, showstringspaces=false, breakindent=0pt]
Therefore, at the leftmost and rightmost points, the weight is experiencing both a centripetal force (due to tension) and a gravitational force (downward). Since these are perpendicular, the resultant force is the combination of the two. However, wait, at the leftmost and rightmost positions, is there a tangential acceleration? Let's see. If the object is moving in a circle with varying speed, then the tangential acceleration is present. At the leftmost point, if the object is moving downward (in a clockwise rotation), then the gravitational force has a component along the direction of motion, which would cause a tangential acceleration. Wait, but at the exact leftmost point, the velocity is purely downward, and gravity is acting straight down. However, the tension at that point is horizontal. So at that exact point, the gravitational force is entirely tangential (since the tangent at the leftmost point is vertical). Therefore, the gravitational force is entirely contributing to the tangential acceleration, while the tension provides the centripetal force. But wait, if that's the case, then the net force would have two components: tension (centripetal) and gravity (tangential). Hence, the resultant force isn't purely towards the center, which is why the problem states that the resultant force does not always point towards the center.

Similarly, at the rightmost point, the velocity is upward (for clockwise motion), and gravity is acting downward. At the exact rightmost point, the velocity is purely upward, tangent to the circle. Gravity is acting downward, which is opposite to the direction of motion. Therefore, gravity here is providing a tangential deceleration, again contributing to a tangential component of acceleration. Tension here is horizontal towards the center (leftward at the rightmost point). Thus, again, the resultant force is a combination of the tension (centripetal) and gravity (tangential), resulting in a net force that is not purely centripetal.

Therefore, at the leftmost and rightmost positions, the resultant force on the weight is the vector sum of the tension force (directed towards the center of the circle) and the gravitational force (directed downward). Since these forces are perpendicular to each other, the resultant force vector will point in a direction that is a combination of centripetal (horizontal) and tangential (vertical) components. This resultant force is not purely towards the center because of the gravitational force acting downward, which introduces a tangential component. Hence, the net force is not centripetal at these points but has both centripetal and tangential components, leading to a resultant force that does not point directly towards the center of the circle.

But wait, let me make sure. Let's take the leftmost point: the tension is to the right (centripetal direction), gravity is down. The net force is the vector sum of these two. So the resultant force would be diagonally right and down. Similarly, at the rightmost point, tension is to the left, gravity is down, so net force is diagonally left and down. So neither of these resultant forces points directly towards the center of the circle. Therefore, the reason the resultant force doesn't always point towards the center is that, in vertical circular motion, gravity acts downward and is not aligned with the string's tension, introducing a tangential component to the net force at positions where the string is horizontal (leftmost and rightmost points). In contrast, at the top and bottom, gravity is either aligned with or opposite to the tension, leading to net forces that are centripetal (if we consider the bottom, where tension upwards minus weight provides centripetal; at the top, tension plus weight if any provides centripetal). Wait, but at the top, if the tension is insufficient, the weight might not complete the circle. But assuming it's in motion, then the net force at top and bottom is still towards the center, but at the sides, the net force is a combination of centripetal and tangential components.

\end{lstlisting}
\end{tcolorbox}
\captionsetup{labelformat=empty}
\label{tab:high_resonign_3}
\end{figure*}

\begin{figure*}[th!]
\centering
\small
\begin{tcolorbox}[colback=green!10!white, %
                  colframe=green!30!white, %
                  width=0.99\linewidth, %
                  arc=4mm, %
                  auto outer arc,
                  sharp corners=north, %
                  before upper={\setlength{\parindent}{0em}} %
                  ]
\begin{lstlisting}[basicstyle=\ttfamily, breaklines=true, showstringspaces=false, breakindent=0pt]
Wait, to summarize: In vertical circular motion, due to gravity, the net force (resultant force) has both centripetal and tangential components except at the top and bottom, where gravity is along the vertical axis. At the leftmost and rightmost positions, the tension is horizontal, gravity is vertical, so their vector sum is not purely horizontal (towards the center), hence the resultant force points towards the center and downward, making it not purely centripetal but a combination.

Therefore, the answer would be that at the leftmost and rightmost positions, the resultant force is a combination of the tension force towards the center and the gravitational force downward. The resultant force does not always point towards the center because gravity introduces a tangential component to the net force at these positions.

I need to make sure there's no error here. Another way to think about it is, in uniform circular motion, the net force must be centripetal, but when the motion is non-uniform (speed changing), there is a tangential component. Since gravity causes the speed to change as the weight moves around the circle, the motion is non-uniform, hence the net force has tangential components. At leftmost and rightmost positions, the tangential component is vertical (due to gravity), so indeed the net force has both centripetal and tangential parts, resulting in a non-centripetal resultant direction.

Yes, that makes sense. Therefore, the conclusion is correct.
</think>

At the leftmost and rightmost positions of the weight's vertical circular motion, the **resultant force** is the vector sum of two forces:  
1. **Tension** in the string, directed horizontally toward the center of the circle.  
2. **Gravity** (weight), acting vertically downward.  

Since these forces are perpendicular to each other, the resultant force vector points diagonally **toward the center and downward** at the leftmost position and diagonally **toward the center and downward** at the rightmost position (due to gravity's consistent downward pull). This resultant force has both a **centripetal component** (from tension) and a **tangential component** (from gravity).  

**Why the resultant force does not always point toward the center:**  
In vertical circular motion, gravity introduces a **tangential acceleration** component at positions where the string is horizontal (leftmost and rightmost points). This tangential component causes the speed of the weight to change (increase on the descending side and decrease on the ascending side), making the motion non-uniform. Consequently, the net force must account for both the centripetal force required for circular motion and the tangential force due to gravity. The resultant force vector therefore does not point purely toward the center but instead combines centripetal and tangential directions.  

At the top and bottom of the circle, gravity aligns with the radial direction, resulting in a purely centripetal net force. However, at the sides (leftmost and rightmost points), gravity acts tangentially, leading to the non-centripetal resultant force."

Reasoning Strategies: ['self-verification', 'backtracking', 'synthesis', 'discussion', 'exploration', 'analysis', 'calculation', 'explanation', 'generalization']
\end{lstlisting}
\end{tcolorbox}
\caption{Reasoning trace with high count of reasoning strategies, and the annotated reasoning strategies.}
\label{tab:high_resonign_4}
\end{figure*}